\pdfoutput=1

\documentclass[11pt]{article}
\usepackage[table,xcdraw]{xcolor}

\usepackage{acl}

\usepackage{times}
\usepackage{latexsym}
\usepackage{enumitem}
\usepackage{algorithm}
\usepackage{algpseudocode}
\usepackage{wrapfig}
\usepackage{booktabs}
\usepackage{multirow}
\usepackage{hyperref}

\usepackage[T1]{fontenc}

\usepackage[utf8]{inputenc}

\usepackage{microtype}

\usepackage{graphicx}
\usepackage{array} 
\usepackage{tabularx}
\usepackage{amssymb}
\usepackage{float}

\title{Defining and Evaluating Visual Language Models' Basic Spatial Abilities: A Perspective from Psychometrics}

\author{
Wenrui Xu\textsuperscript{1*}, Dalin Lyu\textsuperscript{1*}, Weihang Wang\textsuperscript{1*}, Jie Feng\textsuperscript{2}, Chen Gao\textsuperscript{3}, Yong Li\textsuperscript{2}\\
\textsuperscript{1}School of Architecture, Tsinghua University\\
\textsuperscript{2}Department of Electronic Engineering, Tsinghua University\\
\textsuperscript{3}BNRist, Tsinghua University\\
\texttt{\{xwr23,lvdl24,wang-wh22\}@mails.tsinghua.edu.cn}\\
\texttt{\{fengjie,chgao96,liyong07\}@tsinghua.edu.cn}\\
}

\begin{document}

\maketitle

\renewcommand{\thefootnote}{\fnsymbol{footnote}}
\footnotetext[1]{These authors contributed equally to this work.}

\begin{abstract} 


The Theory of Multiple Intelligences underscores the hierarchical nature of cognitive capabilities. To advance Spatial Artificial Intelligence, we pioneer a psychometric framework defining five Basic Spatial Abilities (BSAs) in Visual Language Models (VLMs): \textit{Spatial Perception}, \textit{Spatial Relation}, \textit{Spatial Orientation}, \textit{Mental Rotation}, and \textit{Spatial Visualization}. Benchmarking 13 mainstream VLMs through nine validated psychometric experiments reveals significant gaps versus humans (average score 24.95 vs. 68.38), with three key findings: 1) VLMs mirror human hierarchies (strongest in 2D orientation, weakest in 3D rotation) with independent BSAs (Pearson's r<0.4); 2) Smaller models such as Qwen2-VL-7B surpass larger counterparts, with Qwen leading (30.82) and InternVL2 lagging (19.6); 3) Interventions like chain-of-thought (0.100 accuracy gain) and 5-shot training (0.259 improvement) show limits from architectural constraints. Identified barriers include weak geometry encoding and missing dynamic simulation. By linking psychometric BSAs to VLM capabilities, we provide a diagnostic toolkit for spatial intelligence evaluation, methodological foundations for embodied AI development, and a cognitive science-informed roadmap for achieving human-like spatial intelligence.

\end{abstract}

\renewcommand{\thefootnote}{\arabic{footnote}}
\begin{figure}[t!]

    \centering
    \includegraphics[width=1\linewidth]{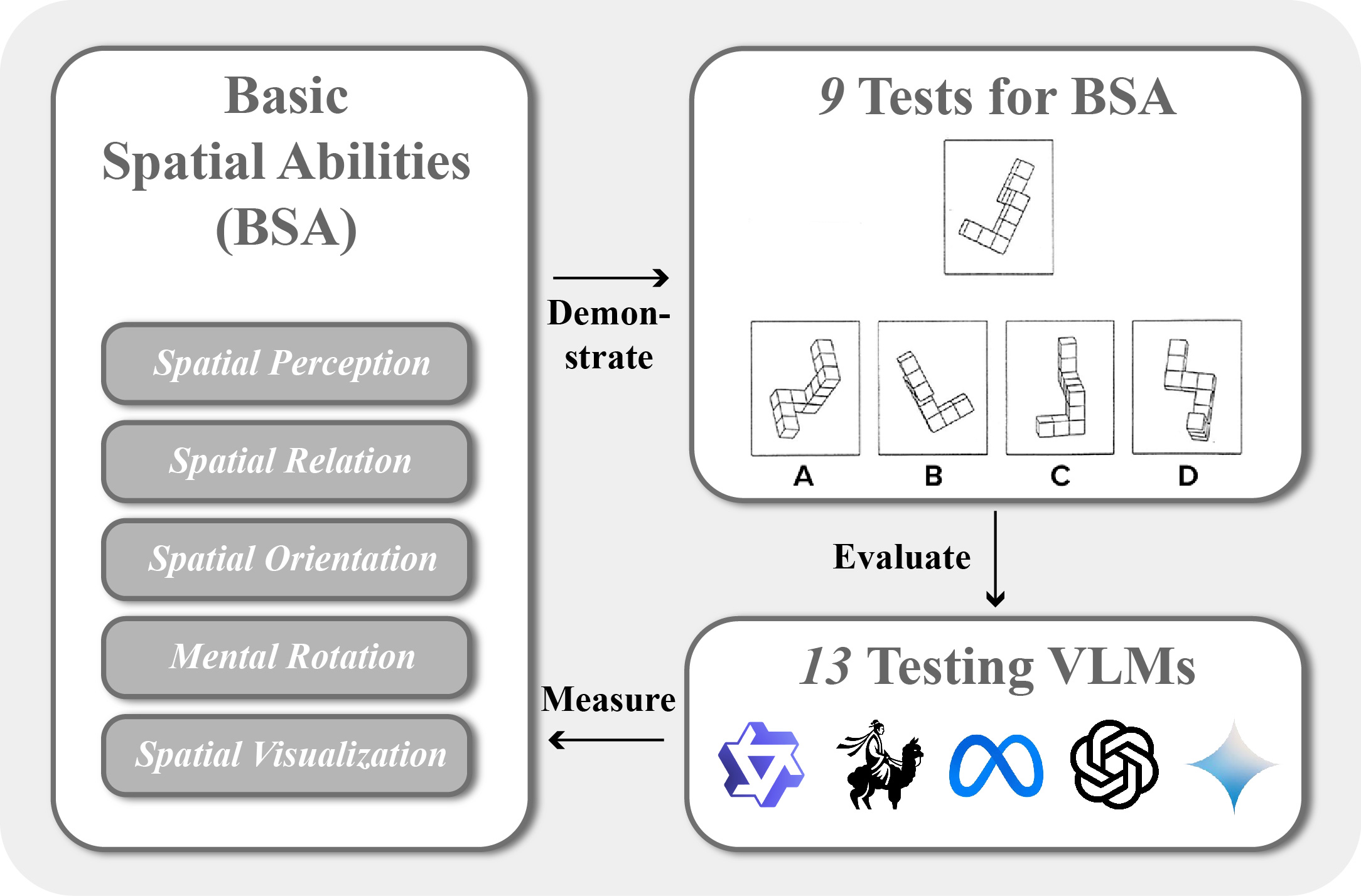}
    \caption{BSA measuring framework for VLMs\footnotemark[1].}
    \label{fig:framework}
    
\end{figure}
\footnotetext[1]{Image source: \citet{vandenbergMentalRotationsGroup1978}.}

\section{Introduction} 

Visual Language Models (VLMs) excel in a wide range of specific tasks \cite{hong3DLLMInjecting3D2023}. However, achieving human-like spatial intelligence for embodied AI applications such as visual navigation and embodied Q\&A remains a challenge \cite{duranteInteractiveAgentFoundation2024,duanSurveyEmbodiedAI2022}. Recent studies reveal that even advanced models like GPT-4o fail basic 2D spatial reasoning tasks that humans solve effortlessly \cite{tangSparkleMasteringBasic2024}.

Gardner's Theory of Multiple Intelligences \cite{bornsteinFramesMindTheory1986}, which is widely accepted across disciplines, posits that human intelligence is hierarchical, with general intelligence (\textit{g}) supported by subordinate intelligences including spatial intelligence. Spatial intelligence itself is also structured hierarchically, including five Basic Spatial Abilities (BSAs): Spatial Perception, Spatial Relation, Spatial Orientation, Mental Rotation, and Spatial Visualization. These abilities are crucial for advanced intelligence and provide a structured framework for evaluation.

However, existing studies assessing the spatial abilities of VLMs lack a solid theoretical foundation and typically focus on isolated abilities without a comprehensive framework, making it challenging to compare results across different studies or uncover potential interconnections between these abilities. Furthermore, most research omits human performance benchmarks, leaving the gap between VLMs and human largely unexplored.

To address these gaps, we propose a psychometric framework using standardized human spatial tests to systematically evaluate VLMs' BSAs, thereby establishing benchmarks and pathways for enhancing spatial intelligence in AI systems.

\section{Related works} 

\subsection{Psychometric Studies on Human Spatial Intelligence}

Human spatial intelligence, defined as the ability to mentally model and manipulate spatial environments \cite{bornsteinFramesMindTheory1986}, has been studied since Spearman first identified it as an independent domain in 1904 \cite{spearmanGeneralIntelligenceObjectively1904}. Research diverges into two complementary streams \cite{poratSpatialAbilityUnderstanding2023}:

\textbf{Psychometric Classification.} The first stream establishes hierarchical intelligence models, from general intelligence to domain-specific intelligence like spatial intelligence \cite{gearySpatialAbilityDistinct2022, spearmanGeneralIntelligenceObjectively1904}. These frameworks enable systematic measurement of spatial abilities through standardized tests, continually refining ability subtypes and assessment methods.

\textbf{Interdisciplinary Mechanisms.} The second stream integrates evolutionary psychology, developmental studies, and cognitive neuroscience to explore the origins and mechanisms of spatial abilities, complementing psychometric taxonomies.

Psychometric consensus defines spatial intelligence through three hierarchical levels \cite{caemmererIndividualIntelligenceTests2020, mcgrewCHCTheoryHuman2009, johnsonStructureHumanIntelligence2005, linnEmergenceCharacterizationSex1985, maccobyPsychologySexDifferences1974, michaelDescriptionSpatialvisualizationAbilities1957, thurstonePrimaryAbilitiesVisual1950}:

\begin{enumerate}

\item General Intelligence (\textit{g}): cross-domain cognitive processes, such as attentional control \cite{kaneRolePrefrontalCortex2002}, neural integration \cite{jungParietofrontalIntegrationTheory2007}, and cellular processes \cite{gearyInclassAttentionSpatial2021}, which impact cross-domain learning and performance.

\item Domain-Specific Intelligence: Abilities in particular domains that share common features, with spatial intelligence representing a distinct set of abilities \cite{vernonAbilityFactorsEnvironmental1965}, as formalized in models such as the CHC theory \cite{carrollHumanCognitiveAbilities1993, hornOrganizationAbilitiesDevelopment1968, cattellTheoryFluidCrystallized1963}.

\item Basic Spatial Abilities (BSAs): Decomposing spatial intelligence into measurable subskills \cite{johnsonSexDifferencesMental2007, hegartySpatialAbilitiesDifferent2006, maierSpatialGeometrySpatial1996, voyerMagnitudeSexDifferences1995, halpernSexDifferencesCognitive1992, pellegrinoUnderstandingSpatialAbility1984}.

\end{enumerate}

This study focuses on Level 3, adopting established human experimental paradigms to evaluate VLMs' basic spatial abilities.

\subsection{Evaluation of LLMs and VLMs' Spatial Abilities}

Recent advances in Visual Language Models (VLMs) have spurred evaluations of their spatial abilities, yet existing studies remain fragmented (Table \ref{tab:existing_studies}). Most prior work focuses on text-based LLMs, assessing abstract spatial relations through verbal descriptions \cite{yamadaEvaluatingSpatialUnderstanding2024}, inherently neglecting visual-spatial processing. Emerging VLM evaluations primarily target specialized 3D tasks (e.g., robotic trajectory labeling \cite{sharmaExploringImprovingSpatial2023}, indoor scene captioning \cite{fuSceneLLMExtendingLanguage2024}), which partially engage Spatial Perception, Spatial Relation, and Spatial Orientation. However, critical gaps persist: 

\begin{enumerate}

\item Theoretical Disconnect: Tasks lack grounding in theoretical frameworks, preventing direct comparison with human cognition. For instance, robotic trajectory tests \cite{sharmaExploringImprovingSpatial2023} conflate spatial reasoning with action planning.
\item Limited Scope: Most studies omit Mental Rotation and Spatial Visualization (Table \ref{tab:existing_studies}, MR/SV columns).
\item Benchmark Absence: No study systematically maps VLM performance to hierarchical BSAs or provides human baselines.

\end{enumerate}

This study addresses these limitations through a psychometric evaluation framework. Unlike prior studies testing subsets of BSAs (e.g., \citet{fuSceneLLMExtendingLanguage2024}: SP+SR+SO), we assess all five BSAs using standardized human experiments, enabling cross-model and human-AI comparisons.

\begin{table}[t]
\centering
\caption{Existing studies testing LLMs and VLMs' spatial abilities.}
\label{tab:existing_studies}
\resizebox{\linewidth}{!}{%
\begin{tabular}{lcccccc}
\hline
\multirow{2}{*}{Related Work} & \multirow{2}{*}{VLM} & \multicolumn{5}{c}{Tested Spatial Abilities} \\ \cline{3-7} 
 &  & SP & SR & SO & MR & SV \\ \hline
\citet{fuSceneLLMExtendingLanguage2024}          & Yes & \checkmark & \checkmark & \checkmark &            &  \\
\citet{tangSparkleMasteringBasic2024}            & Yes & \checkmark & \checkmark & \checkmark &            &  \\
\citet{sharmaExploringImprovingSpatial2023}      & Yes &            & \checkmark & \checkmark &            &  \\
\citet{hong3DLLMInjecting3D2023}                 & Yes & \checkmark & \checkmark & \checkmark &            &  \\
\citet{bangMultitaskMultilingualMultimodal2023}  & Yes &            & \checkmark & \checkmark &            &  \\
\citet{yamadaEvaluatingSpatialUnderstanding2024} & No  &            & \checkmark &            &            &  \\
\citet{momennejadEvaluatingCognitiveMaps2023}    & No  &            & \checkmark & \checkmark &            &  \\
\hyperref[section:bib]{Cohn et al. (2023)}       & No  &            & \checkmark &            & \checkmark &  \\
\hline
This study                                       & Yes & \checkmark & \checkmark & \checkmark & \checkmark & \checkmark \\
\hline
\end{tabular}%
}
\end{table}

\section{Methodology}

\begin{figure}[t]
    \centering
    \includegraphics[width=1\linewidth]{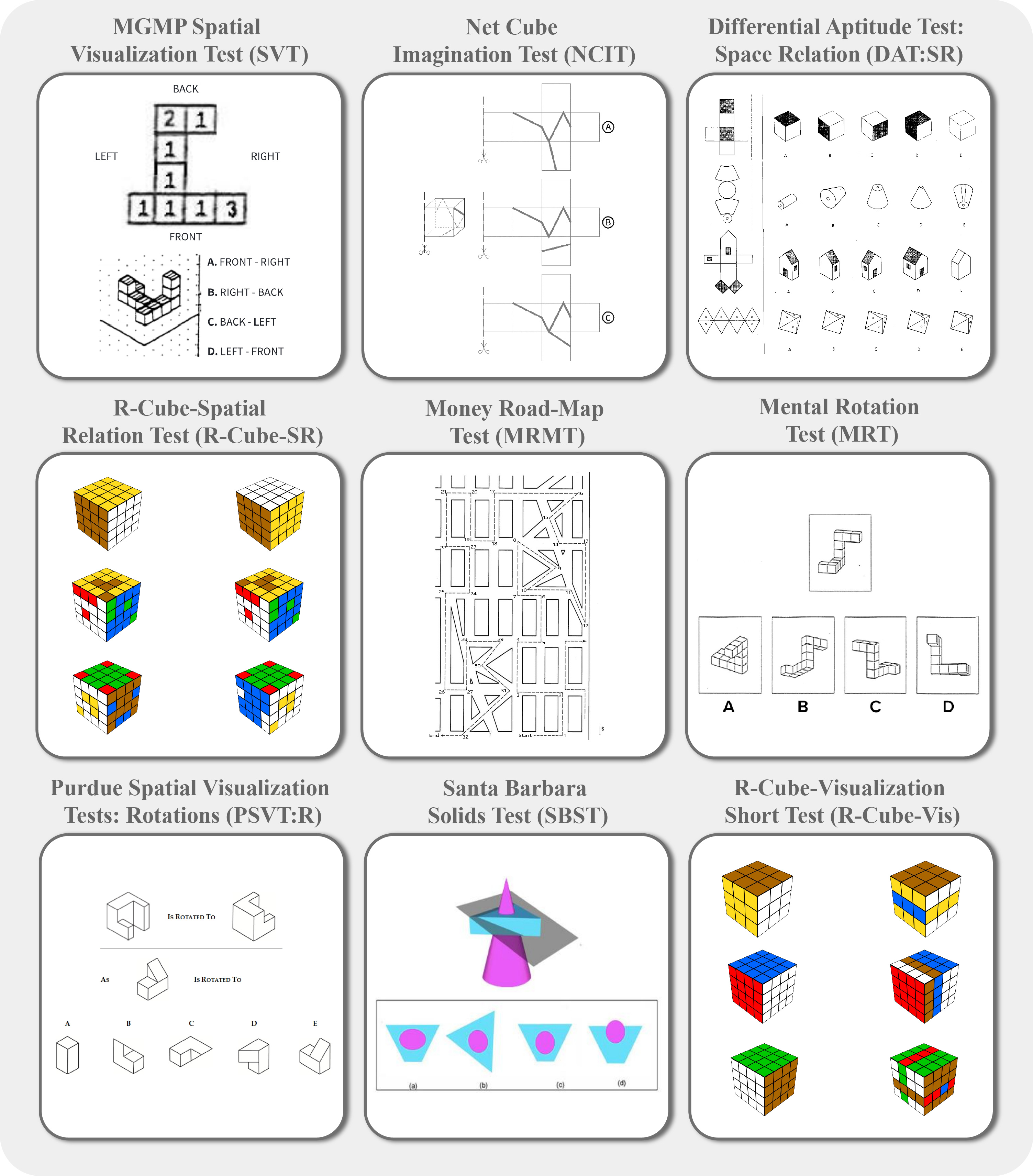}
    \caption{Examples of classic BSA Tests\footnotemark[2].}
    \label{fig:Classic_Tests}
\end{figure}
\footnotetext[2]{Image source: \citet{erkekRelationshipPreserviceTeachers2011, ben-chaimDevelopmentAnalysisSpatial1986, pawlak-jakubowskaEvaluationSTEMStudents2023, katsioloudisComparativeAnalysisSpatial2014, Bennett2012TheDA, fehringerRCubeSRTestNew2023, friedmanComputerizedSpatialOrientation2020, vingerhoetsAnalysisMoneyRoadmap1996, vandenbergMentalRotationsGroup1978, petersRedrawnVandenbergKuse1995, bodnerPurdueVisualizationRotations1997, maedaPsychometricPropertiesRevised2013, hegartyHowSpatialAbilities2009, fehringerSupplementaryMaterialsImplementation2021}.}

\subsection{Definition and Composition of Basic Spatial Abilities}
Based on Psychometric theories \cite{maierSpatialGeometrySpatial1996} and existing VLM studies, this study identified five key dimensions of Basic Spatial Abilities \cite{pawlak-jakubowskaEvaluationSTEMStudents2023}, decomposing the concept as a comprehensive whole, as shown in Table \ref{tab:Decomposed BSA and the corresponding tests}.
To carry out a complete spatial ability evaluation of VLMs, the study selected nine specific classic psychometric tests, as shown in Figure \ref{fig:Classic_Tests}, designed to cover all five aspects of BSAs, enhancing the persuasiveness of the test. Human experiment results from existing studies were included as benchmarks for comparison.

\begin{table}[t]
\caption{Decomposed BSA and the corresponding tests.}
\label{tab:Decomposed BSA and the corresponding tests}
\resizebox{1\linewidth}{!}{
\begin{tabular}{ccc}
\hline
\textbf{Type} &
  \textbf{Definition} &
  \textbf{Tests} \\ \hline
\begin{tabular}[c]{@{}c@{}}Spatial \\ Perception\end{tabular} &
  \begin{tabular}[c]{@{}c@{}}The ability to perceive \\ horizontal and vertical \\ orientations without \\ interference from \\ miscellaneous information.\end{tabular} &
  SVT \\ \hline
\begin{tabular}[c]{@{}c@{}}Spatial \\ Relation\end{tabular} &
  \begin{tabular}[c]{@{}c@{}}The ability of recognizing \\ relationships between \\ parts of an entity.\end{tabular} &
  \begin{tabular}[c]{@{}c@{}}NCIT \\ DAT:SR \\ R-Cube-SR\end{tabular} \\ \hline
\begin{tabular}[c]{@{}c@{}}Spatial \\ Orientation\end{tabular} &
  \begin{tabular}[c]{@{}c@{}}The ability to navigate \\ or enter a given \\ spatial state.\end{tabular} &
  MRMT \\ \hline
\begin{tabular}[c]{@{}c@{}}Mental \\ Rotation\end{tabular} &
  \begin{tabular}[c]{@{}c@{}}The ability to mentally \\ rotate 3D objects.\end{tabular} &
  \begin{tabular}[c]{@{}c@{}}MRT \\ PSVT:R\end{tabular} \\ \hline
\begin{tabular}[c]{@{}c@{}}Spatial \\ Visualization\end{tabular} &
  \begin{tabular}[c]{@{}c@{}}The ability to mentally \\ manipulate and transform \\ 2D and 3D objects.\end{tabular} &
  \begin{tabular}[c]{@{}c@{}}SBST \\ R-Cube-Vis\end{tabular} \\ \hline
\end{tabular}%
}
\end{table}

\subsection{Tests for Basic Spatial Abilities}
\label{section:Tests_for_Basic_Spatial_Abilities}

The goal of the employed tests, on the one hand, is to evaluate the spatial abilities of different VLMs and to compare them with those of humans. On the other hand, by breaking down the basic abilities of spatial intelligence, we aim to identify the deficiencies in current VLMs and provide a necessary foundation for future research aimed at enhancing these abilities. To achieve this goal, the nine selected test questions sets were carefully screened and are all currently available that unambiguously demonstrate a certain BSA, have complete questions, answers, and reproducible human experiment results, which are widely recognized. These questions were not created by the researchers but are grounded in solid psychometric theoretical foundations. The study posits that using as many classic tests as possible can more authentically and accurately reflect the BSAs of VLMs. Therefore, arbitrarily deleting tests to balance the weights of the five core spatial competencies was deemed inappropriate.

Tests adopted are presented in the format of multiple-choice or true/false questions. In each specific test, after briefly explaining the test's content and the ways of answering, we provide the questions in the form of images.

\subsubsection{Spatial Perception Tests}
\textbf{MGMP Spatial Visualization Test (SVT)}. 
SVT, originally developed for the Middle Grades Mathematics Project (MGMP), is also known as the "Lappan Test" and is composed of 32 multiple choice items, each with five options \cite{erkekRelationshipPreserviceTeachers2011,ben-chaimDevelopmentAnalysisSpatial1986}. The test utilizes the combination and transformation of square-cube buildings. The subject is expected to imagine the 2D flat view, the 3D corner view, and the "map plan", which is a numeric cube description of the base of the building. Questions include imagining the conversion between 2D and 3D views, the final appearance when some cubes are altered, and calculating the number of cubes used in a building.

\subsubsection{Spatial Relation Tests}
\textbf{Net Cube Imagination Test (NCIT)}. 
NCIT is based on the conversion between 2D and 3D cubes with lines drawn on inner or outer faces \cite{pawlak-jakubowskaEvaluationSTEMStudents2023}. Each of the 16 tasks has three options. The first eight items require expanding the cube into a flat shape, while the rest involve the reverse development of the cube. 

\textbf{Differential Aptitude Test: Space Relation (DAT:SR)}. 
DAT:SR is part of the Differential Aptitude Test, measuring the ability to relate two and three-dimensional worlds \cite{katsioloudisComparativeAnalysisSpatial2014,Bennett2012TheDA}. The test consists of 40 items and focuses on folding and unfolding 3D geometric shapes. When provided 2D flat unfolded figures, subjects are required to choose the 3D restored form from four options and vice versa.

\textbf{R-Cube-Spatial Relation Test (R-Cube-SR)}. 
R-Cube-SR uses Rubik's cubes with six different colors on each side as rotated visual materials \cite{fehringerRCubeSRTestNew2023}. For each of the 48 items, two cubes are shown in the corner view, the right of which may be the possible rotated result of the left cube. The subjects are required to give a true/false answer with a limited view of the colored sides. The test was created in plain and pattern versions.

\subsubsection{Spatial Orientation Tests}
\textbf{Money Road-Map Test (MRMT)}. 
The Standardized Road-Map Test of Direction Sense \cite{friedmanComputerizedSpatialOrientation2020,vingerhoetsAnalysisMoneyRoadmap1996} commonly known as MRMT, requires allocentric to egocentric right/left discrimination. Subjects follow a dashed path with 32 right/left turns on a map of an abstract city and are required to indicate the direction taken at each turn according to the facing direction. The turn types include the ones that require no rotation, a standard rotation of approximately 90°, and an irregular rotation between 90°-180°.

\subsubsection{Mental Rotation Tests}
\textbf{Mental Rotation Test (MRT)}. 
A classic test developed by \citet{vandenbergMentalRotationsGroup1978} takes the direct way of rotating 3D geometric figures made by cubes. Subjects are required to choose from the four options, two correct rotated reproductions of the target figure. The revised version of 24 items \cite{petersRedrawnVandenbergKuse1995} is utilized.

\textbf{Purdue Spatial Visualization Tests: Visualization of Rotations (PSVT:R)}. 
Developed by \citet{bodnerPurdueVisualizationRotations1997}, PSVT:R is an independent extended version of the Purdue Spatial Visualization Test, which requires matching between the target and the rotated objects. For each item, a geometric shape is rotated first to show the target rotation pattern. Subjects are required to choose from the five rotated views the correct rotated result of another shape under the same pattern. \citet{maedaPsychometricPropertiesRevised2013} further advanced the test to 30 test items, 17 of which involve asymmetrical shapes.

\subsubsection{Spatial Visualization Tests}
\textbf{Santa Barbara Solids Test (SBST)}. 
SBST involves geometric solids intersected by a cutting plane, and the task is to imagine the 2D cross section of the solids and choose from the four options, the two correct answers \cite{hegartyHowSpatialAbilities2009}. Two varying parameters result in the different difficulty of the 30 items: geometric complexity and cutting plane orientation. Simple solids, joined solids, and embedded solids take up one third of the items respectively, while orthogonal and oblique cutting planes divide the items at the same time.

\textbf{R-Cube-Visualization Short Test (R-Cube-Vis)}. 
Similar to the R-Cube-SR test, R-Cube-Vis items consist of two juxtaposed Rubik's cubes \cite{fehringerSupplementaryMaterialsImplementation2021}. However, instead of rotating as a whole, the composing cubes can be rotated as well. Subjects are required to decide the possibility of the left cube rotated into the right one in the 60 items. Based on the size of the cube, the number of the rotated elements, and the ways they are rotated, the items are divided into six difficulty levels.

\section{Experiments}

\begin{figure*}[!ht]
    \centering
    \includegraphics[width=1\linewidth]{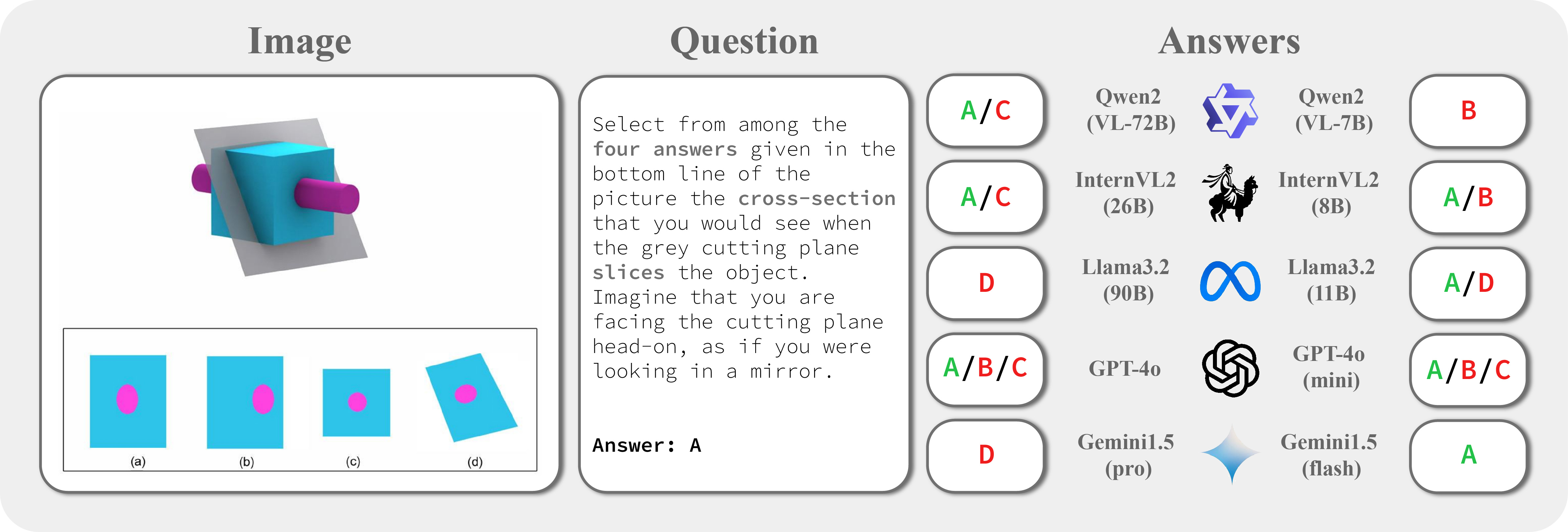}
    \caption{An example of different VLMs' answers to a particular Santa Barbara Solids Test question\footnotemark[3].}
    \label{fig:tests_used}
\end{figure*}

\subsection{Settings}

\textbf{Models and APIs}. 
For BSA evaluation, we tested 13 mainstream open-source and commercial models, showcasing the full spectrum of the current major models' BSAs. For commercial VLMs, we used GPT-4o, GPT-4o mini, and GPT-4 Turbo from OpenAI \cite{yangDawnLMMsPreliminary2023} and Gemini-1.5-pro, Gemini-1.5-flash, and Gemini-1.5-flash-8b from Google \cite{geminiteamGeminiFamilyHighly2024}. For open-source models, we tested Qwen2-VL-72B, Qwen2-VL-7B \cite{baiQwenVLFrontierLarge2023}, InternVL2-Llama3, InternVL2-26B, InternVL2-8B \cite{chenHowFarAre2024}, Llama-3.2-11B, Llama-3.2-90B \cite{llamateamai@metaLlama3Herd2024} (the Qwen2 and Llama models are instruct variant). To carry out large-scale automatic tests, we used APIs from four platforms, including SiliconFlow, DeepInfra, OpenAI, and Google.

\textbf{Data Processing and Evaluation Metrics}. 
A total of 312 questions are fed to each of the models. Among the questions, spatial perception accounts for 10.26\% (32 questions); spatial relation accounts for 33.33\% (104 questions); spatial orientation accounted for 10.26\% (32 questions); mental rotation accounts for 17.31\% (54 questions); and spatial visualization accounts for 28.84\% (90 questions). For the assessment of individual abilities, a greater number of questions can increase the credibility of the results. For the evaluation of overall ability, the study equates different individual abilities, so that no single ability will have a greater weight due to a larger number of questions, nor will it have a biased impact on the overall ability.

Using zero-shot or few-shot experiments, we designed the following prompt for VLMs: "You are taking a spatial ability test. Please output the result in pure text format as: question number, answer." Each question is presented to the VLMs separately, while the question and options are displayed in a single image, because the case study indicates negligible difference of model performance with and without explicit separation of answer choices, and the ability to properly recognize option indices are also considered an elementary ability beneath spatial abilities. To ensure comparability with human benchmarks, we retained the instruction phase from the original human experiments when testing VLMs. The ground truth sets come from the original tests or human analysis.

For all employed tests, we take accuracy as the evaluation metric and consider the answer correct when it matches the ground-truth set exactly, and partly correct when it includes a correct option but does not contain false ones. The score for each question is calculated as the number of correct options selected divided by the total number of correct options. Specifically, scores are not counted to avoid distorted results if the models guess the same answers for all items due to the failure to understand the task.

The maximum score for each test is 100, and the model's score is determined by the number of correct answers divided by the total number of questions. The score for each BSA is calculated as the average score of all tests under that ability, while the Overall Ability Score is the average of the scores for all BSAs.

\footnotetext[3]{Image source: \citet{hegartyHowSpatialAbilities2009}.}

\begin{figure*}[t!]
    \centering
    \includegraphics[width=0.85\linewidth]{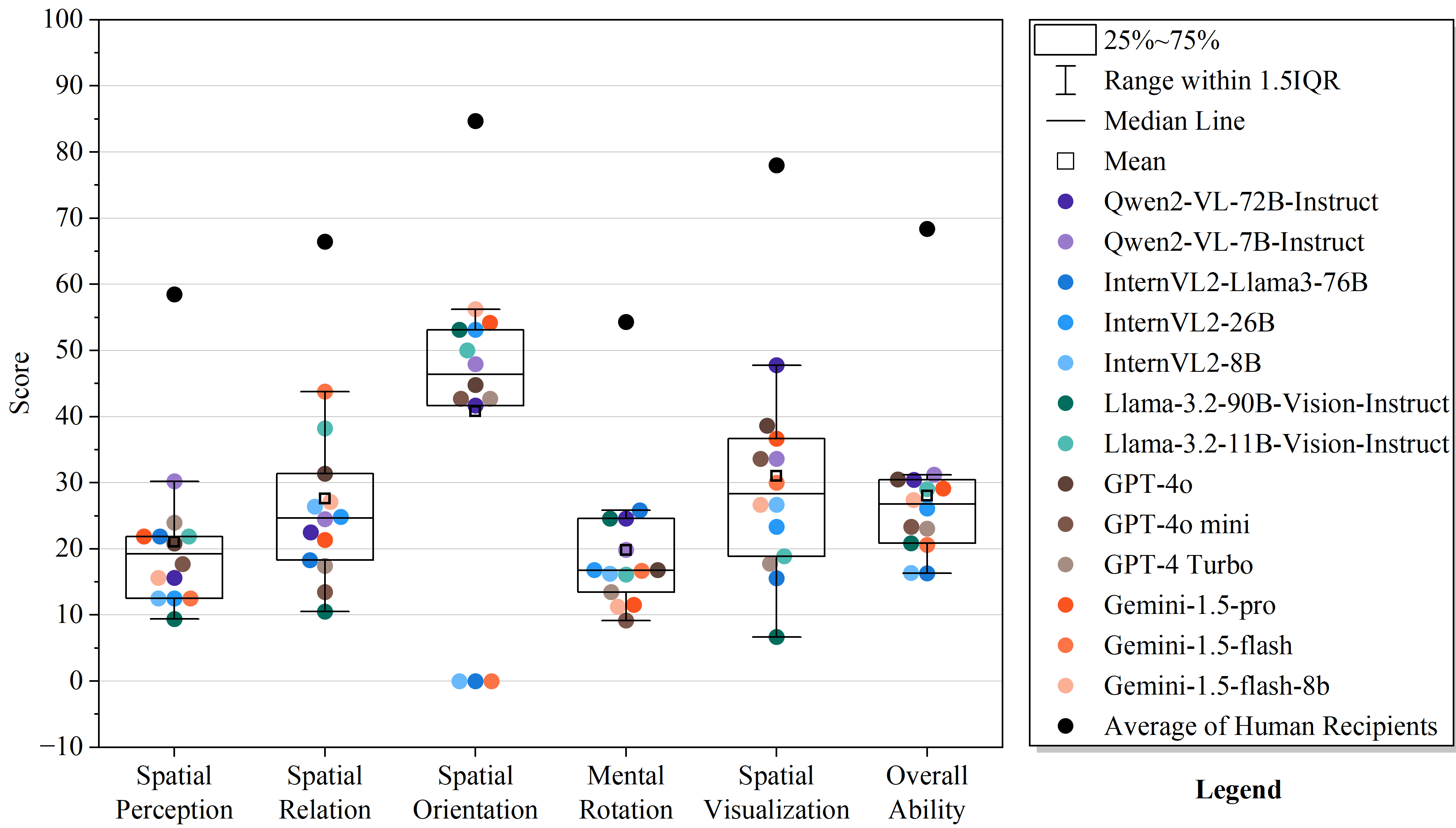}
    \caption{Comparison of VLMs' and humans' five basic spatial abilities and overall ability.}
    \label{fig:ability_results}
\end{figure*}

\subsection{Results}
\label{section:results}
As shown in Table \ref{tab:test_result}, each cell in the table represents a model's score on a specific test. To evaluate the stability of model performance, each test was repeated three times for every model. The numbers in parentheses indicate the standard deviation of these three scores, serving as a measure of performance stability. The final rows of the table present the average scores and standard deviations for all models across each test, along with the corresponding human experiment results.

\subsubsection{Human vs VLM}

As illustrated in Figure \ref{fig:ability_results}, the overall ability scores of the 13 VLMs are relatively close, ranging from 16.31 to 31.22, with an average of 24.95. This is significantly lower than the human average of 68.38. When analyzed across the five BSAs, human performance consistently surpasses that of VLMs. In terms of averages, both human and VLMs show the same performance ranking across the abilities, with spatial orientation being the best and mental rotation the worst.

\subsubsection{VLM vs VLM}

From Figure \ref{fig:ability_results}, it can be observed that the performance of VLMs show consistency in spatial perception and mental rotation. In contrast, spatial relation and spatial visualization exhibit more significant variability. Most models perform relatively well in spatial orientation, yet the discrepancy is also considerable, with three models (Intern-VL2-76B, Intern-VL2-8B, and Gemini-1.5-flash) failing completely. In terms of overall ability, the performance differences among models are not particularly significant, though some models (e.g. Qwen2-VL-7B and GPT-4o), stand out with relatively strong results. 

\subsubsection{Correlation of Basic Spatial Abilities}

\begin{figure}[t!]
    \centering
    \includegraphics[width=1\linewidth]{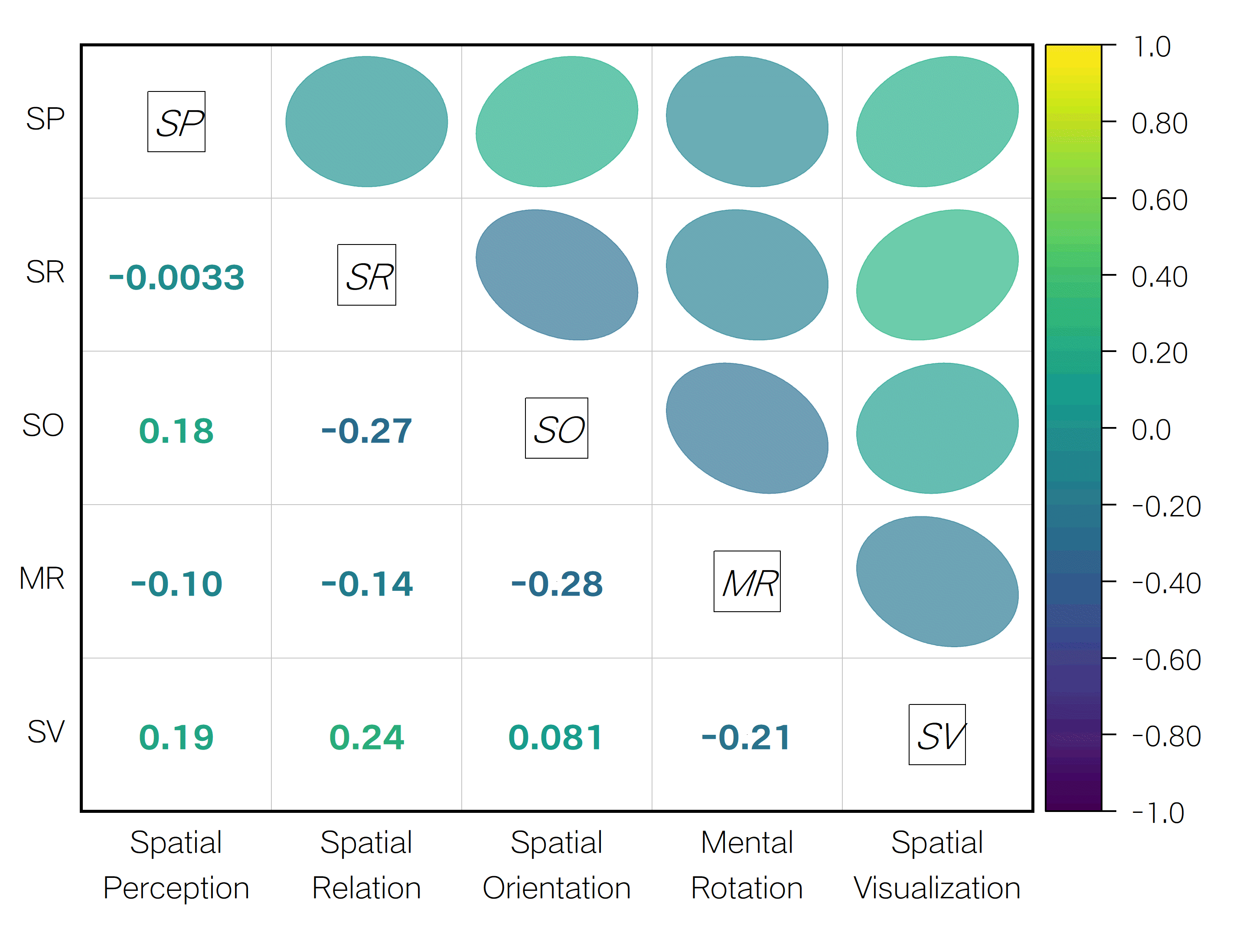}
    \caption{Pearson's correlation coefficients (r) between five basic spatial abilities.}
    \label{fig:Correlation of BSAs.}
\end{figure}

For the test results of the models' BSAs, we performed a correlation analysis between the abilities to examine their associations. We used the Pearson correlation test, as shown in Figure \ref{fig:Correlation of BSAs.}. The results show that Pearson's correlation coefficients (r) between any two variables are less than 0.4, indicating that all of the ability combinations show very weak or no correlation. Since the events satisfied the two-dimensional normal distribution, no correlation is equivalent to independence. Thus, it can be considered that the spatial abilities do not show a correlation with each other, meaning that each spatial ability is sufficiently "basic" and proves the credibility of the BSA framework.

\begin{figure}[t!]
    \centering
    \includegraphics[width=1\linewidth]{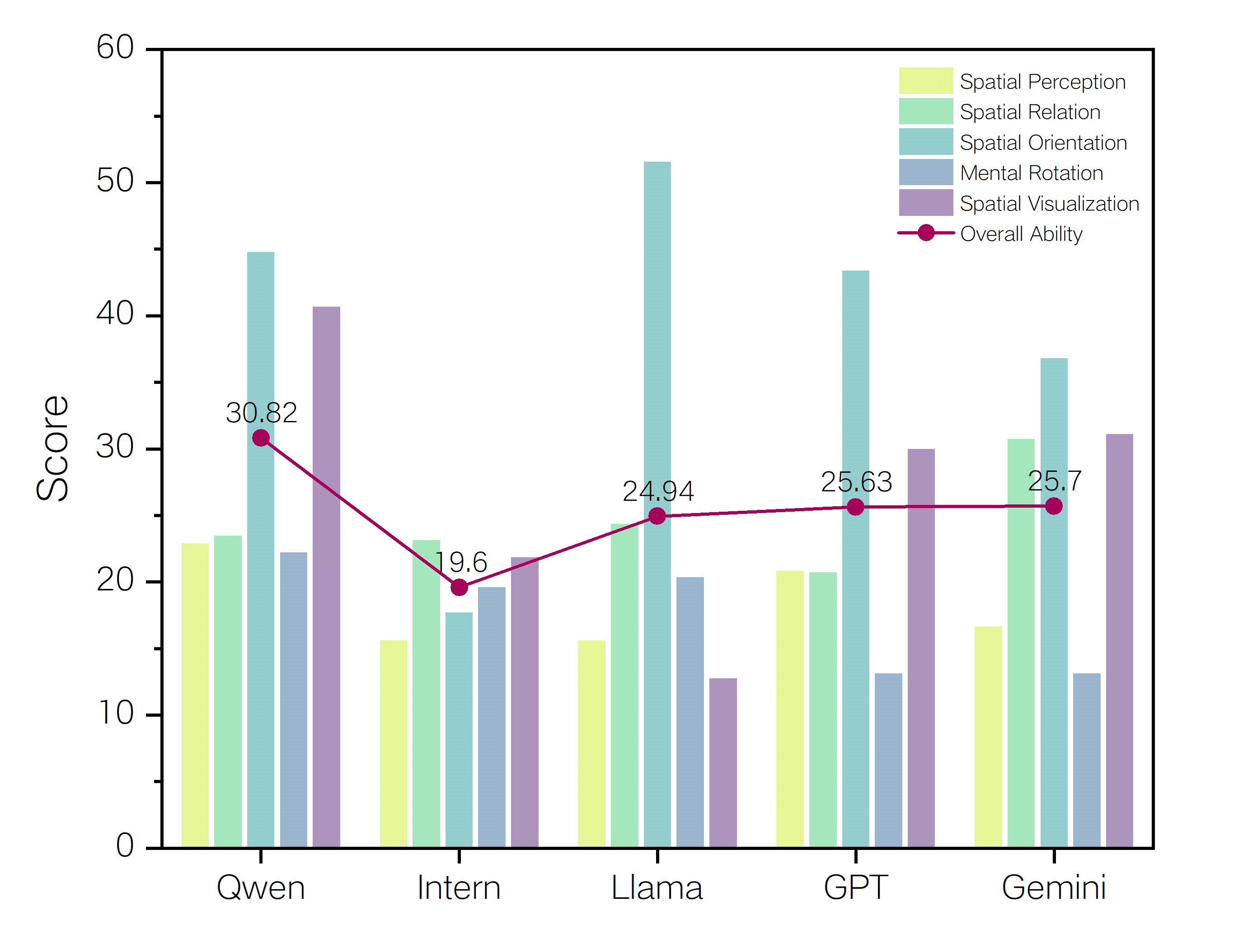}
    \caption{Comparison of different series of VLMs' basic spatial abilities and overall ability.}
    \label{fig:manufacturer}
\end{figure}

\begin{figure*}[t!]
    \centering
    \includegraphics[width=1\linewidth]{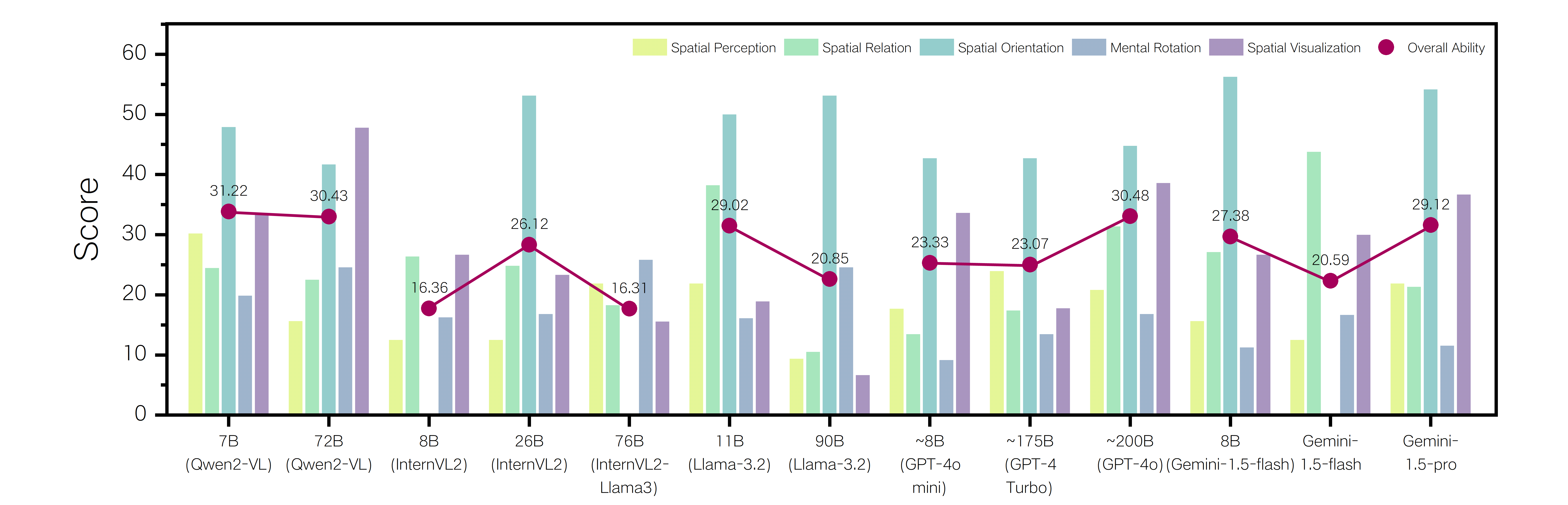}
    \caption{Comparison of the basic spatial abilities of VLMs with different sizes and manufacturers.}
    \label{fig:model parameters}
\end{figure*}

\subsection{Discussion}
\subsubsection{VLMs' Overall Performance}

Our evaluation confirms a significant gap between VLMs’ basic spatial abilities and human benchmarks across all dimensions, aligning with prior studies on individual abilities (e.g., spatial relations \cite{yamadaEvaluatingSpatialUnderstanding2024, cohnDialecticalLanguageModel2023}, spatial orientation \cite{momennejadEvaluatingCognitiveMaps2023}). This raises fundamental questions about whether VLMs operate as programmable pattern recognizers or genuinely emulate human-like spatial intelligence.

Notably, VLMs mirror human performance rankings across BSAs: highest in spatial orientation, followed by spatial visualization, and lowest in mental rotation (Figure \ref{fig:ability_results}). This correlates with task complexity that 2D tasks require simpler coordinate reasoning, while 3D demands dynamic mental manipulation, suggesting VLMs struggle disproportionately with higher-dimensional transformations just like humans. This performance gradient implies a developmental pathway: strengthening basic 2D spatial reasoning may scaffold advanced 3D capabilities \cite{tangSparkleMasteringBasic2024}. Such hierarchical training strategies could bridge current limitations.

\subsubsection{Impact of VLM Manufacturer and Size}

As shown in Figure \ref{fig:manufacturer}, models from different manufacturers exhibit noticeable performance differences. The Qwen series demonstrates clear superiority with an overall score of 30.82, outperforming competitors across multiple assessment dimensions. Mid-tier performers including Gemini, GPT, and Llama series cluster around 25, while InternVL2 trails notably at 19.6.

Notably, individual spatial ability performance shows distinct patterns from aggregate scores. Gemini-1.5-flash achieves peak spatial relation performance (43.78) but demonstrates marked deficiencies in mental rotation (16.67) and complete absence of spatial orientation capability. Conversely, the Llama series excels in spatial orientation while underperforming in spatial visualization. These disparities likely originate from heterogeneous training data distributions across different BSAs.

Turning to model size analysis (Figure \ref{fig:model parameters}), our findings challenge conventional expectations. Contrary to typical scaling laws, we observe no positive correlation between parameter count and BSA performance. Smaller models (leftward positioned in manufacturer groupings) frequently outperform their larger counterparts, a trend particularly evident in the Qwen and Llama series where around 10B-parameter models surpass larger variants. The Gemini-1.5-flash-8B matches the performance of its substantially larger Gemini-1.5-pro counterpart, with similar scaling anomalies observed in InternVL2 models. This suggests that architectural optimization and training methodologies may outweigh sheer parameter count for spatial reasoning tasks, with compact models offering favorable efficiency-performance tradeoffs.

To clarify, model parameter counts for Qwen, InternVL2, and Llama series derive from official specifications, while GPT series estimates follow Microsoft's disclosed information. Gemini series parameters remain undisclosed, though relative comparisons within the series remain valid.

\subsubsection{Impact of CoT and Example-Based Training}

To assess intervention strategies, we implemented a five-step chain-of-thought (CoT) protocol for spatial reasoning (understanding the 3D shape, analyzing the plane, determining the cross-section, matching cross-section to options, giving the answer, as shown in Figure \ref{fig:CoT}) and conducted example-based training on the SBST geometric cutting test.

\begin{figure}[t!]
    \centering
    \includegraphics[width=1\linewidth]{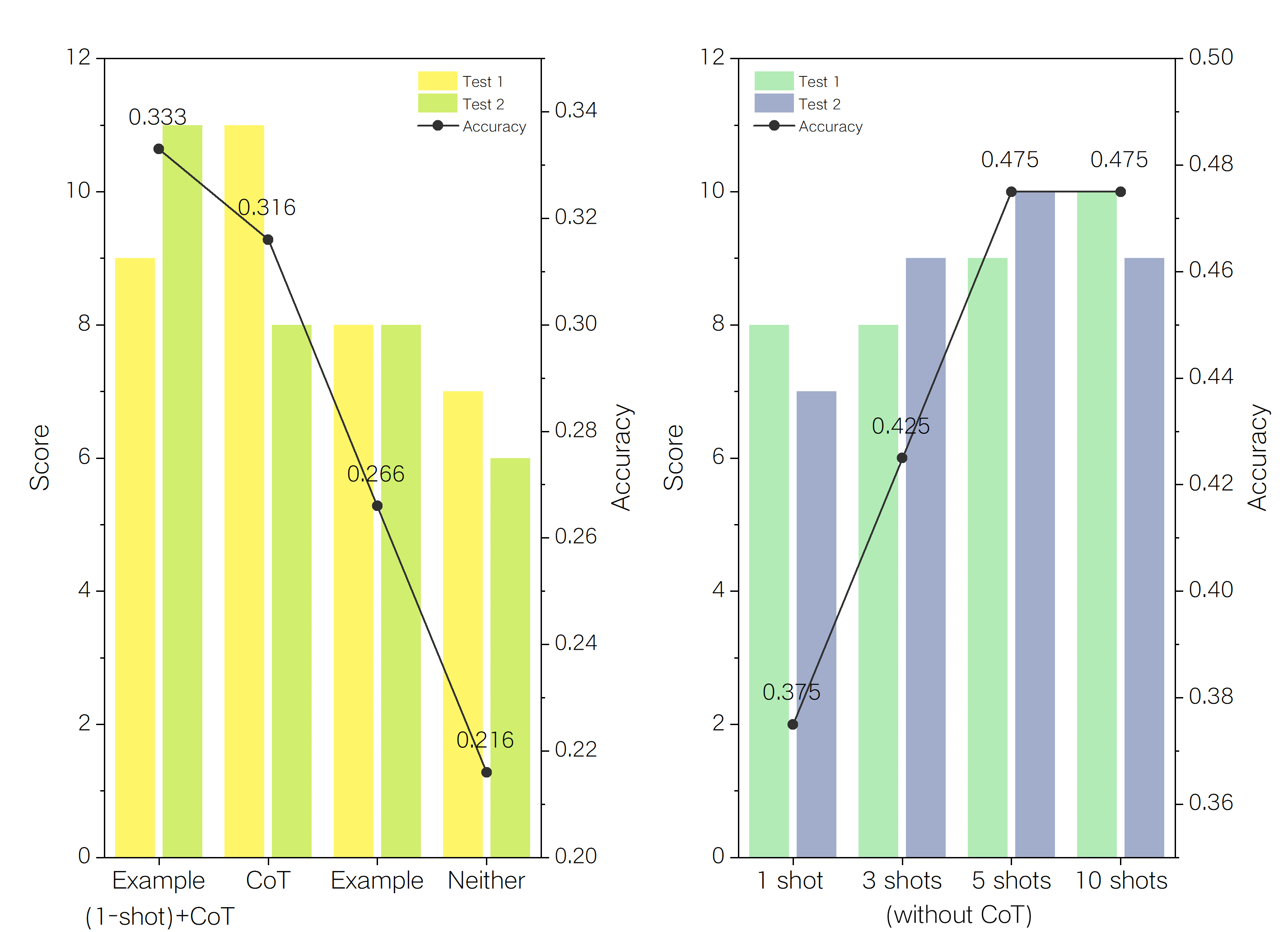}
    \caption{The impact of CoT method and example-based training on the accuracy of VLMs' answers.}
    \label{fig:Training Accuracy}
\end{figure}

As shown in Figure \ref{fig:Training Accuracy}, implementing CoT alone boosted baseline accuracy by 0.100, demonstrating its capacity to enhance visual-semantic alignment and multi-step reasoning. When combined with one-shot example, accuracy further increased by 0.017, suggesting synergistic effects between structured reasoning and pattern recognition.

Additionally, incremental example exposure from one to ten shots initially improved accuracy by 0.159 at 1 shot and 0.209 at 3 shots, but gains plateaued beyond 5 examples (Figure \ref{fig:Training Accuracy}). While examples help VLMs recognize recurring spatial patterns such as prism cross-sections, they fail to address fundamental limitations in dynamic 3D mental simulation, as VLMs overly relied on the general patterns learnt during pre-training process.

\subsubsection{Constraints in VLMs' Basic Spatial Abilities}

Our analysis reveals four fundamental constraints in VLMs’ BSAs. Firstly, VLMs struggle to distinguish subtle shape variations such as hexagon/octagon cross-sections and misinterpret spatial relationships such as interior/exterior boundaries, indicating weak metric encoding in visual representations. This limitation may stem from constraints in training data or a lack of prior knowledge such as geometric principles, which example-based training can partially mitigate. Secondly, even with CoT prompting, VLMs exhibit shallow reasoning chains and cannot dynamically simulate 3D transformations such as mental rotation trajectories, contrasting human parietal lobe-driven simulation mechanisms \cite{jungParietofrontalIntegrationTheory2007}. Thirdly, erratic behaviors like multi-answer selection in single-choice 3D tasks expose poor cross-modal grounding, which is a critical gap between textual instruction parsing and visual feature extraction. Lastly, overreliance on pre-training patterns limits adaptability to novel spatial configurations.

These limitations underscore architectural deficiencies beyond data scarcity. Unlike humans who integrate dorsal and ventral visual streams for spatial processing, VLMs lack dedicated modules for dynamic spatial simulation. Bridging this gap may require hybrid architectures embedding geometric priors and neurosymbolic reasoning.



\section{Conclusion}

This study establishes a psychometric framework for evaluating five basic spatial abilities (BSAs) in visual language models (VLMs), benchmarking 13 mainstream models across nine rigorously designed experiments. Our findings reveal three key insights:

First, VLMs exhibit a significant performance gap compared to humans (average score: 24.95 vs. 68.38) while paradoxically mirroring human performance hierarchies across BSAs, excelling in 2D spatial orientation but struggling with 3D mental rotation. The statistical independence of these BSAs validates our framework's discriminative capacity.

Second, model performance varies significantly by manufacturer, reflecting architectural priorities. Qwen-series models demonstrate superior cross-BSA integration, while InternVL2's fragmented performance suggests biases in training data. Contrary to scaling laws, compact models like Qwen2-VL-7B and Gemini-1.5-flash-8B frequently outperform larger counterparts, suggesting that spatial reasoning depends more on architectural optimization than parameter size, with profound implications for efficient model design.

Third, intervention strategies yield measurable but limited improvements. Chain-of-thought (CoT) prompting enhances visual-semantic alignment, improving baseline accuracy by 0.100, while 5-shot training boosts accuracy by 0.259 through pattern recognition. However, these methods plateau at certain limits that VLMs fail to simulate dynamic 3D transformations even with CoT training, and struggle with metric encoding, highlighting architectural constraints beyond training data limitations.

Additionally, our analysis identifies four critical barriers to spatial intelligence: 1) Weak geometric priors in visual representations; 2) Lack of dynamic simulation mechanisms akin to human parietal processing; 3) Disjointed cross-modal grounding; 4) Over-reliance on static pre-training patterns. Overcoming these may require a paradigm shift toward hybrid architectures integrating neurosymbolic reasoning and biologically inspired spatial processing.

In conclusion, by establishing the first quantitative linkage between psychometric BSAs and VLM capabilities, this study provides both a diagnostic toolkit for evaluating spatial intelligence and a roadmap for advancing embodied AI systems. Future progress depends on close collaboration between machine learning and cognitive science communities.




\section*{Ethics Statement}

This study utilized the API service provided by OpenAl, DeepInfra, Google, and SiliconFlow in full compliance with their terms of service and usage policies.

\bibliography{anthology,custom,xwr_references}

\begin{thebibliography}{53}
\expandafter\ifx\csname natexlab\endcsname\relax\def\natexlab#1{#1}\fi

\bibitem[{Bai et~al.(2023)Bai, Bai, Yang, Wang, Tan, Wang, Lin, Zhou, and Zhou}]{baiQwenVLFrontierLarge2023}
Jinze Bai, Shuai Bai, Shusheng Yang, Shijie Wang, Sinan Tan, Peng Wang, Junyang Lin, Chang Zhou, and Jingren Zhou. 2023.
\newblock \href {https://doi.org/10.48550/arXiv.2308.12966} {Qwen-{{VL}}: A frontier large vision-language model with versatile abilities}.

\bibitem[{Bang et~al.(2023)Bang, Cahyawijaya, Lee, Dai, Su, Wilie, Lovenia, Ji, Yu, Chung, Do, Xu, and Fung}]{bangMultitaskMultilingualMultimodal2023}
Yejin Bang, Samuel Cahyawijaya, Nayeon Lee, Wenliang Dai, Dan Su, Bryan Wilie, Holy Lovenia, Ziwei Ji, Tiezheng Yu, Willy Chung, Quyet~V. Do, Yan Xu, and Pascale Fung. 2023.
\newblock \href {https://doi.org/10.48550/arXiv.2302.04023} {A multitask, multilingual, multimodal evaluation of {{ChatGPT}} on reasoning, hallucination, and interactivity}.

\bibitem[{{Ben-Chaim} et~al.(1986){Ben-Chaim}, Lappan, and Houang}]{ben-chaimDevelopmentAnalysisSpatial1986}
D.~{Ben-Chaim}, G.~Lappan, and R.~T. Houang. 1986.
\newblock \href {https://doi.org/10.2466/pms.1986.63.2.659} {Development and analysis of a spatial visualization test for middle school boys and girls}.
\newblock \emph{Perceptual and Motor Skills}, 63(2 Pt 1):659--669.

\bibitem[{Bennett et~al.(2012)Bennett, Seashore, and Wesman}]{Bennett2012TheDA}
George~K. Bennett, Harold~G. Seashore, and Alexander~G. Wesman. 2012.
\newblock The differential aptitude test : A review and critique.

\bibitem[{Bodner and Guay(1997)}]{bodnerPurdueVisualizationRotations1997}
George Bodner and Roland Guay. 1997.
\newblock \href {https://doi.org/10.1007/s00897970138a} {The purdue visualization of rotations test}.
\newblock \emph{Chemical Educator}, 2:1--17.

\bibitem[{Bornstein and Gardner(1986)}]{bornsteinFramesMindTheory1986}
Marc~H. Bornstein and Howard Gardner. 1986.
\newblock \href {https://doi.org/10.2307/3332707} {Frames of mind: The theory of multiple intelligences}.
\newblock \emph{Journal of Aesthetic Education}, 20(2):120.

\bibitem[{Caemmerer et~al.(2020)Caemmerer, Keith, and Reynolds}]{caemmererIndividualIntelligenceTests2020}
Jacqueline~M. Caemmerer, Timothy~Z. Keith, and Matthew~R. Reynolds. 2020.
\newblock \href {https://doi.org/10.1016/j.intell.2020.101433} {Beyond individual intelligence tests: Application of cattell-horn-carroll theory}.
\newblock \emph{Intelligence}, 79:101433.

\bibitem[{Carroll(1993)}]{carrollHumanCognitiveAbilities1993}
John~B. Carroll. 1993.
\newblock \href {https://doi.org/10.1017/CBO9780511571312} {\emph{Human {{Cognitive Abilities}}: {{A Survey}} of {{Factor-Analytic Studies}}}}.
\newblock Cambridge University Press, Cambridge.

\bibitem[{Cattell(1963)}]{cattellTheoryFluidCrystallized1963}
Raymond~B. Cattell. 1963.
\newblock \href {https://doi.org/10.1037/h0046743} {Theory of fluid and crystallized intelligence: A critical experiment}.
\newblock \emph{Journal of Educational Psychology}, 54(1):1--22.

\bibitem[{Chen et~al.(2024)Chen, Wang, Tian, Ye, Gao, Cui, Tong, Hu, Luo, Ma, Ma, Wang, Dong, Yan, Guo, He, Shi, Jin, Xu, Wang, Wei, Li, Zhang, Zhang, Cai, Wen, Yan, Dou, Lu, Zhu, Lu, Lin, Qiao, Dai, and Wang}]{chenHowFarAre2024}
Zhe Chen, Weiyun Wang, Hao Tian, Shenglong Ye, Zhangwei Gao, Erfei Cui, Wenwen Tong, Kongzhi Hu, Jiapeng Luo, Zheng Ma, Ji~Ma, Jiaqi Wang, Xiaoyi Dong, Hang Yan, Hewei Guo, Conghui He, Botian Shi, Zhenjiang Jin, Chao Xu, Bin Wang, Xingjian Wei, Wei Li, Wenjian Zhang, Bo~Zhang, Pinlong Cai, Licheng Wen, Xiangchao Yan, Min Dou, Lewei Lu, Xizhou Zhu, Tong Lu, Dahua Lin, Yu~Qiao, Jifeng Dai, and Wenhai Wang. 2024.
\newblock \href {https://doi.org/10.48550/arXiv.2404.16821} {How far are we to {{GPT-4V}}? {{Closing}} the gap to commercial multimodal models with open-source suites}.

\bibitem[{Cohn and {Hernandez-Orallo}(2023)}]{cohnDialecticalLanguageModel2023}
Anthony~G. Cohn and Jose {Hernandez-Orallo}. 2023.
\newblock \href {https://doi.org/10.48550/arXiv.2304.11164} {Dialectical language model evaluation: An initial appraisal of the commonsense spatial reasoning abilities of {{LLMs}}}.

\bibitem[{Duan et~al.(2022)Duan, Yu, Tan, Zhu, and Tan}]{duanSurveyEmbodiedAI2022}
Jiafei Duan, Samson Yu, Hui~Li Tan, Hongyuan Zhu, and Cheston Tan. 2022.
\newblock \href {https://doi.org/10.1109/TETCI.2022.3141105} {A survey of embodied {{AI}}: From simulators to research tasks}.
\newblock \emph{IEEE Transactions on Emerging Topics in Computational Intelligence}, 6(2):230--244.

\bibitem[{Durante et~al.(2024)Durante, Sarkar, Gong, Taori, Noda, Tang, Adeli, Lakshmikanth, Schulman, Milstein, Terzopoulos, Famoti, Kuno, Llorens, Vo, Ikeuchi, {Fei-Fei}, Gao, Wake, and Huang}]{duranteInteractiveAgentFoundation2024}
Zane Durante, Bidipta Sarkar, Ran Gong, Rohan Taori, Yusuke Noda, Paul Tang, Ehsan Adeli, Shrinidhi~Kowshika Lakshmikanth, Kevin Schulman, Arnold Milstein, Demetri Terzopoulos, Ade Famoti, Noboru Kuno, Ashley Llorens, Hoi Vo, Katsu Ikeuchi, Li~{Fei-Fei}, Jianfeng Gao, Naoki Wake, and Qiuyuan Huang. 2024.
\newblock \href {https://doi.org/10.48550/arXiv.2402.05929} {An interactive agent foundation model}.

\bibitem[{Erkek et~al.(2011)Erkek, Isiksal, and Cakiroglu}]{erkekRelationshipPreserviceTeachers2011}
{\"O}zlem Erkek, Mine Isiksal, and Erdinc Cakiroglu. 2011.
\newblock \emph{The Relationship between Preservice Teachers' Spatial Anxiety and Geometry Self-Efficacy in Terms of Gender and Undergraduate Program}.

\bibitem[{Fehringer(2021)}]{fehringerSupplementaryMaterialsImplementation2021}
Benedict Fehringer. 2021.
\newblock Supplementary materials for: Implementation of the {{R-cube-vis}} test in its long and short version in english as well as german.

\bibitem[{Fehringer(2023)}]{fehringerRCubeSRTestNew2023}
Benedict C. O.~F. Fehringer. 2023.
\newblock \href {https://doi.org/10.1027/1015-5759/a000682} {R-{{Cube-SR Test}}: {{A New Test}} for {{Spatial Relations Distinguishable From Visualization}}}.
\newblock \emph{European Journal of Psychological Assessment}, 39(1):37--48.

\bibitem[{Friedman et~al.(2020)Friedman, Kohler, Gunalp, Boone, and Hegarty}]{friedmanComputerizedSpatialOrientation2020}
Alinda Friedman, Bernd Kohler, Peri Gunalp, Alexander~P. Boone, and Mary Hegarty. 2020.
\newblock \href {https://doi.org/10.3758/s13428-019-01277-3} {A computerized spatial orientation test}.
\newblock \emph{Behavior Research Methods}, 52(2):799--812.

\bibitem[{Fu et~al.(2024)Fu, Liu, Chen, Nie, and Xiong}]{fuSceneLLMExtendingLanguage2024}
Rao Fu, Jingyu Liu, Xilun Chen, Yixin Nie, and Wenhan Xiong. 2024.
\newblock \href {https://doi.org/10.48550/arXiv.2403.11401} {Scene-{{LLM}}: {{Extending Language Model}} for {{3D Visual Understanding}} and {{Reasoning}}}.

\bibitem[{Geary(2022)}]{gearySpatialAbilityDistinct2022}
David~C. Geary. 2022.
\newblock \href {https://doi.org/10.1016/j.intell.2021.101616} {Spatial ability as a distinct domain of human cognition: An evolutionary perspective}.
\newblock \emph{Intelligence}, 90:101616.

\bibitem[{Geary et~al.(2021)Geary, Hoard, Nugent, and Scofield}]{gearyInclassAttentionSpatial2021}
David~C. Geary, Mary~K. Hoard, Lara Nugent, and John~E. Scofield. 2021.
\newblock \href {https://doi.org/10.1037/edu0000487} {In-class attention, spatial ability, and mathematics anxiety predict across-grade gains in adolescents' mathematics achievement}.
\newblock \emph{Journal of Educational Psychology}, 113(4):754--769.

\bibitem[{{Gemini Team}(2024)}]{geminiteamGeminiFamilyHighly2024}
{Gemini Team}. 2024.
\newblock Gemini: A family of highly capable multimodal models.

\bibitem[{Halpern(1992)}]{halpernSexDifferencesCognitive1992}
Diane~F. Halpern. 1992.
\newblock \emph{Sex Differences in Cognitive Abilities, 2nd Ed}.
\newblock Sex Differences in Cognitive Abilities, 2nd Ed. Lawrence Erlbaum Associates, Inc, Hillsdale, NJ, US.

\bibitem[{Hegarty et~al.(2009)Hegarty, Keehner, Khooshabeh, and Montello}]{hegartyHowSpatialAbilities2009}
Mary Hegarty, Madeleine Keehner, Peter Khooshabeh, and Daniel~R. Montello. 2009.
\newblock \href {https://doi.org/10.1016/j.lindif.2008.04.006} {How spatial abilities enhance, and are enhanced by, dental education}.
\newblock \emph{Learning and Individual Differences}, 19(1):61--70.

\bibitem[{Hegarty et~al.(2006)Hegarty, Montello, Richardson, Ishikawa, and Lovelace}]{hegartySpatialAbilitiesDifferent2006}
Mary Hegarty, Daniel~R. Montello, Anthony~E. Richardson, Toru Ishikawa, and Kristin Lovelace. 2006.
\newblock \href {https://doi.org/10.1016/j.intell.2005.09.005} {Spatial abilities at different scales: {{Individual}} differences in aptitude-test performance and spatial-layout learning}.
\newblock \emph{Intelligence}, 34(2):151--176.

\bibitem[{Hong et~al.(2023)Hong, Zhen, Chen, Zheng, Du, Chen, and Gan}]{hong3DLLMInjecting3D2023}
Yining Hong, Haoyu Zhen, Peihao Chen, Shuhong Zheng, Yilun Du, Zhenfang Chen, and Chuang Gan. 2023.
\newblock \href {https://doi.org/10.48550/arXiv.2307.12981} {{{3D-LLM}}: {{Injecting}} the {{3D World}} into {{Large Language Models}}}.

\bibitem[{Horn(1968)}]{hornOrganizationAbilitiesDevelopment1968}
John~L. Horn. 1968.
\newblock \href {https://doi.org/10.1037/h0025662} {Organization of abilities and the development of intelligence}.
\newblock \emph{Psychological Review}, 75(3):242--259.

\bibitem[{Johnson and Bouchardjr(2005)}]{johnsonStructureHumanIntelligence2005}
W~Johnson and T~Bouchardjr. 2005.
\newblock \href {https://doi.org/10.1016/j.intell.2004.12.002} {The structure of human intelligence: It is verbal, perceptual, and image rotation ({{VPR}}), not fluid and crystallized}.
\newblock \emph{Intelligence}, 33(4):393--416.

\bibitem[{Johnson and Bouchard(2007)}]{johnsonSexDifferencesMental2007}
Wendy Johnson and Thomas~J. Bouchard. 2007.
\newblock \href {https://doi.org/10.1016/j.intell.2006.03.012} {Sex differences in mental abilities: g masks the dimensions on which they lie}.
\newblock \emph{Intelligence}, 35(1):23--39.

\bibitem[{Jung and Haier(2007)}]{jungParietofrontalIntegrationTheory2007}
Rex~E. Jung and Richard~J. Haier. 2007.
\newblock \href {https://doi.org/10.1017/S0140525X07001185} {The parieto-frontal integration theory ({{P-FIT}}) of intelligence: Converging neuroimaging evidence}.
\newblock \emph{Behavioral and Brain Sciences}, 30(2):135--154.

\bibitem[{Kane and Engle(2002)}]{kaneRolePrefrontalCortex2002}
Michael~J. Kane and Randall~W. Engle. 2002.
\newblock \href {https://doi.org/10.3758/BF03196323} {The role of prefrontal cortex in working-memory capacity, executive attention, and general fluid intelligence: An individual-differences perspective}.
\newblock \emph{Psychonomic Bulletin \& Review}, 9(4):637--671.

\bibitem[{Katsioloudis et~al.(2014)Katsioloudis, Jovanovic, and Jones}]{katsioloudisComparativeAnalysisSpatial2014}
Petros Katsioloudis, Vukica Jovanovic, and Mildred Jones. 2014.
\newblock \href {https://doi.org/10.21061/jte.v26i1.a.6} {A comparative analysis of spatial visualization ability and drafting models for industrial and technology education students}.
\newblock \emph{Journal of Technology Education}, 26:88--104.

\bibitem[{Linn and Petersen(1985)}]{linnEmergenceCharacterizationSex1985}
Marcia~C. Linn and Anne~C. Petersen. 1985.
\newblock \href {https://doi.org/10.2307/1130467} {Emergence and characterization of sex differences in spatial ability: A meta-analysis}.
\newblock \emph{Child Development}, 56(6):1479--1498.

\bibitem[{{Llama Team, AI @ Meta}(2024)}]{llamateamai@metaLlama3Herd2024}
{Llama Team, AI @ Meta}. 2024.
\newblock The llama 3 herd of models.

\bibitem[{Maccoby and Jacklin(1974)}]{maccobyPsychologySexDifferences1974}
Eleanor~E. Maccoby and Carol~N. Jacklin. 1974.
\newblock \emph{The Psychology of Sex Differences}.
\newblock The Psychology of Sex Differences. Stanford University Press.

\bibitem[{Maeda et~al.(2013)Maeda, Yoon, {Kim-Kang}, and Imbrie}]{maedaPsychometricPropertiesRevised2013}
Yukiko Maeda, So~Yoon Yoon, K.~{Kim-Kang}, and P.K. Imbrie. 2013.
\newblock Psychometric properties of the revised {{PSVT}}: {{R}} for measuring first year engineering students' spatial ability.
\newblock \emph{International Journal of Engineering Education}, 29.

\bibitem[{Maier(1996)}]{maierSpatialGeometrySpatial1996}
Peter~Herbert Maier. 1996.
\newblock Spatial geometry and spatial ability--{{How}} to make solid geometry solid.
\newblock In \emph{Selected Papers from the {{Annual Conference}} of {{Didactics}} of {{Mathematics}}}, Osnabrueck, Germany: Gesellschaft f{\"u}r Didaktik der Mathematik (GDM).

\bibitem[{McGrew(2009)}]{mcgrewCHCTheoryHuman2009}
Kevin~S. McGrew. 2009.
\newblock \href {https://doi.org/10.1016/j.intell.2008.08.004} {{{CHC}} theory and the human cognitive abilities project: Standing on the shoulders of the giants of psychometric intelligence research}.
\newblock \emph{Intelligence}, 37(1):1--10.

\bibitem[{Michael et~al.(1957)Michael, Guilford, Fruchter, and Zimmerman}]{michaelDescriptionSpatialvisualizationAbilities1957}
William~B. Michael, J.P. Guilford, Benjamin Fruchter, and Wayne~S. Zimmerman. 1957.
\newblock \href {https://doi.org/10.1177/001316445701700202} {The description of spatial-visualization abilities}.
\newblock \emph{Educational and Psychological Measurement}, 17(2):185--199.

\bibitem[{Momennejad et~al.(2023)Momennejad, Hasanbeig, Vieira, Sharma, Ness, Jojic, Palangi, and Larson}]{momennejadEvaluatingCognitiveMaps2023}
Ida Momennejad, Hosein Hasanbeig, Felipe Vieira, Hiteshi Sharma, Robert~Osazuwa Ness, Nebojsa Jojic, Hamid Palangi, and Jonathan Larson. 2023.
\newblock Evaluating cognitive maps and planning in large language models with {{CogEval}}.
\newblock https://arxiv.org/abs/2309.15129v1.

\bibitem[{{Pawlak-Jakubowska} and Terczy{\'n}ska(2023)}]{pawlak-jakubowskaEvaluationSTEMStudents2023}
Anita {Pawlak-Jakubowska} and Ewa Terczy{\'n}ska. 2023.
\newblock \href {https://doi.org/10.1038/s41598-023-44371-5} {Evaluation of {{STEM}} students' spatial abilities based on a novel net cube imagination test}.
\newblock \emph{Scientific Reports}, 13(1):17296.

\bibitem[{Pellegrino et~al.(1984)Pellegrino, Alderton, and Shute}]{pellegrinoUnderstandingSpatialAbility1984}
James~W. Pellegrino, David~L. Alderton, and Valerie~J. Shute. 1984.
\newblock \href {https://doi.org/10.1080/00461528409529300} {Understanding spatial ability}.
\newblock \emph{Educational Psychologist}, 19(4):239--253.

\bibitem[{Peters et~al.(1995)Peters, Laeng, Latham, Jackson, Zaiyouna, and Richardson}]{petersRedrawnVandenbergKuse1995}
M.~Peters, B.~Laeng, K.~Latham, M.~Jackson, R.~Zaiyouna, and C.~Richardson. 1995.
\newblock \href {https://doi.org/10.1006/brcg.1995.1032} {A {{Redrawn Vandenberg}} and {{Kuse Mental Rotations Test}} - {{Different Versions}} and {{Factors That Affect Performance}}}.
\newblock \emph{Brain and Cognition}, 28(1):39--58.

\bibitem[{Porat and Ceobanu(2023)}]{poratSpatialAbilityUnderstanding2023}
Ronen Porat and Ciprian Ceobanu. 2023.
\newblock \href {https://doi.org/10.15405/epes.23056.9} {Spatial ability: Understanding the past, looking into the future}.
\newblock \emph{European Proceedings of Educational Sciences}, Education, Reflection, Development - ERD 2022.

\bibitem[{Sharma(2023)}]{sharmaExploringImprovingSpatial2023}
Manasi Sharma. 2023.
\newblock \href {https://doi.org/10.48550/arXiv.2312.01054} {Exploring and improving the spatial reasoning abilities of large language models}.

\bibitem[{Spearman(1904)}]{spearmanGeneralIntelligenceObjectively1904}
C.~Spearman. 1904.
\newblock \href {https://doi.org/10.2307/1412107} {"general intelligence," objectively determined and measured}.
\newblock \emph{American Journal of Psychology}, 15(2):201.

\bibitem[{Tang et~al.(2024)Tang, Qu, Wang, Zhuang, Wu, Ma, Wang, Zheng, Zhao, and Zhao}]{tangSparkleMasteringBasic2024}
Yihong Tang, Ao~Qu, Zhaokai Wang, Dingyi Zhuang, Zhaofeng Wu, Wei Ma, Shenhao Wang, Yunhan Zheng, Zhan Zhao, and Jinhua Zhao. 2024.
\newblock \href {https://doi.org/10.48550/arXiv.2410.16162} {\emph{Sparkle: Mastering Basic Spatial Capabilities in Vision Language Models Elicits Generalization to Composite Spatial Reasoning}}.

\bibitem[{Thurstone(1950)}]{thurstonePrimaryAbilitiesVisual1950}
L.~L. Thurstone. 1950.
\newblock \href {http://arxiv.org/abs/3143593} {Some primary abilities in visual thinking}.
\newblock \emph{Proceedings of The American Philosophical Society}, 94(6):517--521.

\bibitem[{Vandenberg and Kuse(1978)}]{vandenbergMentalRotationsGroup1978}
Steven~G. Vandenberg and Allan~R. Kuse. 1978.
\newblock \href {https://doi.org/10.2466/pms.1978.47.2.599} {Mental {{Rotations}}, a {{Group Test}} of {{Three-Dimensional Spatial Visualization}}}.
\newblock \emph{Perceptual and Motor Skills}, 47(2):599--604.

\bibitem[{Vernon(1965)}]{vernonAbilityFactorsEnvironmental1965}
Philip~E. Vernon. 1965.
\newblock \href {https://doi.org/10.1037/h0021472} {Ability factors and environmental influences}.
\newblock \emph{American Psychologist}, 20(9):723--733.

\bibitem[{Vingerhoets et~al.(1996)Vingerhoets, Lannoo, and Bauwens}]{vingerhoetsAnalysisMoneyRoadmap1996}
Guy Vingerhoets, Engelien Lannoo, and Sabien Bauwens. 1996.
\newblock \href {https://doi.org/10.1016/0887-6177(95)00055-0} {Analysis of the money road-map test performance in normal and brain-damaged subjects}.
\newblock \emph{Archives of Clinical Neuropsychology}, 11(1):1--9.

\bibitem[{Voyer et~al.(1995)Voyer, Voyer, and Bryden}]{voyerMagnitudeSexDifferences1995}
Daniel Voyer, Susan Voyer, and M.~Philip Bryden. 1995.
\newblock \href {https://doi.org/10.1037/0033-2909.117.2.250} {Magnitude of sex differences in spatial abilities: {{A}} meta-analysis and consideration of critical variables.}
\newblock \emph{Psychological Bulletin}, 117(2):250--270.

\bibitem[{Yamada et~al.(2024)Yamada, Bao, Lampinen, Kasai, and Yildirim}]{yamadaEvaluatingSpatialUnderstanding2024}
Yutaro Yamada, Yihan Bao, Andrew~K. Lampinen, Jungo Kasai, and Ilker Yildirim. 2024.
\newblock \href {https://doi.org/10.48550/arXiv.2310.14540} {Evaluating {{Spatial Understanding}} of {{Large Language Models}}}.

\bibitem[{Yang et~al.(2023)Yang, Li, Lin, Wang, Lin, Liu, and Wang}]{yangDawnLMMsPreliminary2023}
Zhengyuan Yang, Linjie Li, Kevin Lin, Jianfeng Wang, Chung-Ching Lin, Zicheng Liu, and Lijuan Wang. 2023.
\newblock \href {https://doi.org/10.48550/arXiv.2309.17421} {The dawn of {{LMMs}}: Preliminary explorations with {{GPT-4V}}(ision)}.

\end{thebibliography}
\label{section:bib}
\bibliographystyle{acl_natbib}

\appendix
\onecolumn

\section{Data Statistics}
As a necessary complement to the results in Section \ref{section:results}, this section provides the original statistics of the data from VLMs' BSA evaluation. The complete results are listed in Table \ref{tab:test_result}.

\begin{table}[H]
\centering
\caption{Test results of 13 VLMs and human recipients.}
\label{tab:test_result}
\resizebox{1\linewidth}{!}{%
\begin{tabular}{
>{\columncolor[HTML]{FFFFFF}}l 
>{\columncolor[HTML]{FFFFFF}}l 
>{\columncolor[HTML]{FFFFFF}}l 
>{\columncolor[HTML]{FFFFFF}}l 
>{\columncolor[HTML]{FFFFFF}}l 
>{\columncolor[HTML]{FFFFFF}}l 
>{\columncolor[HTML]{FFFFFF}}l 
>{\columncolor[HTML]{FFFFFF}}l 
>{\columncolor[HTML]{FFFFFF}}l 
>{\columncolor[HTML]{FFFFFF}}l }

\toprule[1pt]

\textbf{} &
  \textbf{SVT} &
  \textbf{NCIT} &
  \textbf{DATSR} &
  \textbf{RCSR} &
  \textbf{MRMT} &
  \textbf{MRT} &
  \textbf{PSVTR} &
  \textbf{SBST} &
  \textbf{RCVis} \\

\midrule[0.5pt]

{\color[HTML]{4527A0} \textbf{Qwen2-VL-72B-Instruct}} &
  \begin{tabular}[c]{@{}l@{}}15.63\\ (0)\end{tabular} &
  \begin{tabular}[c]{@{}l@{}}18.75\\ (0)\end{tabular} &
  \begin{tabular}[c]{@{}l@{}}10.57\\ (1.67)\end{tabular} &
  \begin{tabular}[c]{@{}l@{}}38.19\\ (0.98)\end{tabular} &
  \begin{tabular}[c]{@{}l@{}}41.67\\ (1.47)\end{tabular} &
  \begin{tabular}[c]{@{}l@{}}29.17\\ (0)\end{tabular} &
  \begin{tabular}[c]{@{}l@{}}20.00\\ (2.72)\end{tabular} &
  \begin{tabular}[c]{@{}l@{}}47.78\\ (1.57)\end{tabular} &
  / \\

\midrule[0.5pt]

{\color[HTML]{9575CD} \textbf{Qwen2-VL-7B-Instruct}} &
  \begin{tabular}[c]{@{}l@{}}30.21\\ (1.47)\end{tabular} &
  \begin{tabular}[c]{@{}l@{}}37.50\\ (0)\end{tabular} &
  \begin{tabular}[c]{@{}l@{}}11.46\\ (0.75)\end{tabular} &
  / &
  \begin{tabular}[c]{@{}l@{}}47.92\\ (1.47)\end{tabular} &
  \begin{tabular}[c]{@{}l@{}}20.83\\ (0)\end{tabular} &
  \begin{tabular}[c]{@{}l@{}}18.89\\ (1.57)\end{tabular} &
  \begin{tabular}[c]{@{}l@{}}25.56\\ (1.57)\end{tabular} &
  \begin{tabular}[c]{@{}l@{}}41.67\\ (0)\end{tabular} \\
\midrule[0.5pt]

{\color[HTML]{1976D2} \textbf{InternVL2-Llama3-76B}} &
  \begin{tabular}[c]{@{}l@{}}21.88\\ (0)\end{tabular} &
  \begin{tabular}[c]{@{}l@{}}18.75\\ (0)\end{tabular} &
  \begin{tabular}[c]{@{}l@{}}17.82\\ (1.19)\end{tabular} &
  / &
  / &
  \begin{tabular}[c]{@{}l@{}}25.00\\ (0)\end{tabular} &
  \begin{tabular}[c]{@{}l@{}}26.67\\ (0)\end{tabular} &
  \begin{tabular}[c]{@{}l@{}}15.56\\ (1.57)\end{tabular} &
  / \\
\midrule[0.5pt]

{\color[HTML]{2196F3} \textbf{InternVL2-26B}} &
  \begin{tabular}[c]{@{}l@{}}12.50\\ (0)\end{tabular} &
  \begin{tabular}[c]{@{}l@{}}37.50\\ (0)\end{tabular} &
  \begin{tabular}[c]{@{}l@{}}12.15\\ (1.18)\end{tabular} &
  / &
  \begin{tabular}[c]{@{}l@{}}53.13\\ (0)\end{tabular} &
  \begin{tabular}[c]{@{}l@{}}12.50\\ (3.40)\end{tabular} &
  \begin{tabular}[c]{@{}l@{}}21.11\\ (1.57)\end{tabular} &
  \begin{tabular}[c]{@{}l@{}}23.33\\ (7.20)\end{tabular} &
  / \\
\midrule[0.5pt]

{\color[HTML]{64B5F6} \textbf{InternVL2-8B}} &
  \begin{tabular}[c]{@{}l@{}}12.50\\ (0)\end{tabular} &
  \begin{tabular}[c]{@{}l@{}}43.75\\ (0)\end{tabular} &
  \begin{tabular}[c]{@{}l@{}}9.00\\ (0)\end{tabular} &
  / &
  / &
  \begin{tabular}[c]{@{}l@{}}12.50\\ (0)\end{tabular} &
  \begin{tabular}[c]{@{}l@{}}20.00\\ (0)\end{tabular} &
  \begin{tabular}[c]{@{}l@{}}26.67\\ (2.72)\end{tabular} &
  / \\
\midrule[0.5pt]

{\color[HTML]{00695C} \textbf{Llama-3.2-90B-Vision-Instruct}} &
  \begin{tabular}[c]{@{}l@{}}9.38\\ (0)\end{tabular} &
  \begin{tabular}[c]{@{}l@{}}12.50\\ (0)\end{tabular} &
  \begin{tabular}[c]{@{}l@{}}8.50\\ (0)\end{tabular} &
  / &
  \begin{tabular}[c]{@{}l@{}}53.13\\ (0)\end{tabular} &
  \begin{tabular}[c]{@{}l@{}}29.17\\ (0)\end{tabular} &
  \begin{tabular}[c]{@{}l@{}}20.00\\ (0)\end{tabular} &
  \begin{tabular}[c]{@{}l@{}}6.67\\ (0)\end{tabular} &
  / \\
\midrule[0.5pt]

{\color[HTML]{4DB6AC} \textbf{Llama-3.2-11B-Vision-Instruct}} &
  \begin{tabular}[c]{@{}l@{}}21.88\\ (0)\end{tabular} &
  \begin{tabular}[c]{@{}l@{}}50.00\\ (0)\end{tabular} &
  \begin{tabular}[c]{@{}l@{}}26.45\\ (0)\end{tabular} &
  / &
  \begin{tabular}[c]{@{}l@{}}50.00\\ (0)\end{tabular} &
  \begin{tabular}[c]{@{}l@{}}22.22\\ (1.96)\end{tabular} &
  \begin{tabular}[c]{@{}l@{}}10.00\\ (0)\end{tabular} &
  \begin{tabular}[c]{@{}l@{}}18.89\\ (1.57)\end{tabular} &
  / \\
\midrule[0.5pt]

{\color[HTML]{5D4037} \textbf{GPT-4o}} &
  \begin{tabular}[c]{@{}l@{}}20.83\\ (5.31)\end{tabular} &
  \begin{tabular}[c]{@{}l@{}}37.50\\ (5.10)\end{tabular} &
  \begin{tabular}[c]{@{}l@{}}10.10\\ (3.12)\end{tabular} &
  \begin{tabular}[c]{@{}l@{}}46.53\\ (1.96)\end{tabular} &
  \begin{tabular}[c]{@{}l@{}}44.79\\ (8.20)\end{tabular} &
  \begin{tabular}[c]{@{}l@{}}12.50\\ (5.89)\end{tabular} &
  \begin{tabular}[c]{@{}l@{}}21.11\\ (5.67)\end{tabular} &
  \begin{tabular}[c]{@{}l@{}}36.67\\ (7.20)\end{tabular} &
  \begin{tabular}[c]{@{}l@{}}40.56\\ (2.83)\end{tabular} \\
\midrule[0.5pt]

{\color[HTML]{795548} \textbf{GPT-4o mini}} &
  \begin{tabular}[c]{@{}l@{}}17.71\\ (7.80)\end{tabular} &
  \begin{tabular}[c]{@{}l@{}}18.75\\ (10.21)\end{tabular} &
  \begin{tabular}[c]{@{}l@{}}8.19\\ (1.09)\end{tabular} &
  / &
  \begin{tabular}[c]{@{}l@{}}42.71\\ (1.47)\end{tabular} &
  \begin{tabular}[c]{@{}l@{}}2.78\\ (1.96)\end{tabular} &
  \begin{tabular}[c]{@{}l@{}}15.56\\ (3.14)\end{tabular} &
  \begin{tabular}[c]{@{}l@{}}21.11\\ (1.57)\end{tabular} &
  \begin{tabular}[c]{@{}l@{}}46.11\\ (1.57)\end{tabular} \\
\midrule[0.5pt]

{\color[HTML]{A1887F} \textbf{GPT-4 Turbo}} &
  \begin{tabular}[c]{@{}l@{}}23.96\\ (3.90)\end{tabular} &
  \begin{tabular}[c]{@{}l@{}}18.75\\ (0)\end{tabular} &
  \begin{tabular}[c]{@{}l@{}}16.08\\ (3.78)\end{tabular} &
  / &
  \begin{tabular}[c]{@{}l@{}}42.71\\ (1.47)\end{tabular} &
  \begin{tabular}[c]{@{}l@{}}12.50\\ (3.40)\end{tabular} &
  \begin{tabular}[c]{@{}l@{}}14.44\\ (6.85)\end{tabular} &
  \begin{tabular}[c]{@{}l@{}}17.78\\ (4.16)\end{tabular} &
  / \\
\midrule[0.5pt]

{\color[HTML]{F4511E} \textbf{Gemini-1.5-pro}} &
  \begin{tabular}[c]{@{}l@{}}21.88\\ (0)\end{tabular} &
  \begin{tabular}[c]{@{}l@{}}18.75\\ (0)\end{tabular} &
  \begin{tabular}[c]{@{}l@{}}23.93\\ (0.98)\end{tabular} &
  / &
  \begin{tabular}[c]{@{}l@{}}54.17\\ (2.95)\end{tabular} &
  \begin{tabular}[c]{@{}l@{}}4.17\\ (0)\end{tabular} &
  \begin{tabular}[c]{@{}l@{}}18.89\\ (1.57)\end{tabular} &
  \begin{tabular}[c]{@{}l@{}}36.67\\ (0)\end{tabular} &
  \begin{tabular}[c]{@{}l@{}}40.00\\ (1.36)\end{tabular} \\
\midrule[0.5pt]

{\color[HTML]{FF7043} \textbf{Gemini-1.5-flash}} &
  \begin{tabular}[c]{@{}l@{}}12.50\\ (0)\end{tabular} &
  \begin{tabular}[c]{@{}l@{}}62.50\\ (0)\end{tabular} &
  \begin{tabular}[c]{@{}l@{}}25.05\\ (0)\end{tabular} &
  / &
  / &
  \begin{tabular}[c]{@{}l@{}}16.67\\ (0)\end{tabular} &
  \begin{tabular}[c]{@{}l@{}}16.67\\ (0)\end{tabular} &
  \begin{tabular}[c]{@{}l@{}}30.00\\ (0)\end{tabular} &
  / \\
\midrule[0.5pt]

{\color[HTML]{FFAB91} \textbf{Gemini-1.5-flash-8B}} &
  \begin{tabular}[c]{@{}l@{}}15.63\\ (0)\end{tabular} &
  \begin{tabular}[c]{@{}l@{}}37.50\\ (0)\end{tabular} &
  \begin{tabular}[c]{@{}l@{}}16.72\\ (0.59)\end{tabular} &
  / &
  \begin{tabular}[c]{@{}l@{}}56.25\\ (0)\end{tabular} &
  \begin{tabular}[c]{@{}l@{}}12.50\\ (0)\end{tabular} &
  \begin{tabular}[c]{@{}l@{}}10.00\\ (0)\end{tabular} &
  \begin{tabular}[c]{@{}l@{}}26.67\\ (0)\end{tabular} &
  / \\
\midrule[0.5pt]

\textit{\textbf{Average of 13 VLMs}} &
  \textit{18.19} &
  \textit{31.37} &
  \textit{15.08} &
  \textit{5.90} &
  \textit{37.42} &
  \textit{16.35} &
  \textit{17.95} &
  \textit{25.64} &
  \textit{12.95} \\
\midrule[0.5pt]

Standard Deviation of 13 VLMs &
  5.87 &
  15.19 &
  6.54 &
  16.00 &
  21.84 &
  8.49 &
  4.62 &
  10.60 &
  20.26 \\
\midrule[0.5pt]

\textit{\textbf{Average of human recipients}} &
  \textit{58.47} &
  \textit{55.00} &
  \textit{55.33} &
  \textit{89.00} &
  \textit{84.69} &
  \textit{45.00} &
  \textit{63.60} &
  \textit{68.00} &
  \textit{88.00} \\
\midrule[0.5pt]

Standard Deviation of human &
  19.72 &
  19.13 &
  / &
  9.00 &
  14.41 &
  20.83 &
  20.53 &
  23.00 &
  7.00 \\
\midrule[0.5pt]

Number of human recipients&
  1007 &
  105 &
  1480 &
  51 &
  61 &
  636 &
  1022 &
  223 &
  52 \\
\bottomrule[1pt]
\end{tabular}
}
\end{table}

\begin{table}[H]
\centering
\caption{VLMs' performance in Santa Barbara Solids Test with and without explicit separation of answer choices.}
\label{tab:option_results}
\resizebox{0.8\linewidth}{!}{%
\begin{tabular}{
>{\columncolor[HTML]{FFFFFF}}c 
>{\columncolor[HTML]{FFFFFF}}c 
>{\columncolor[HTML]{FFFFFF}}c }
\toprule[1pt]
\textbf{Model} & \textbf{Original Experiment} & \textbf{Supplementary Experiment} \\ \midrule[0.5pt]
Qwen2-VL-72B-Instruct & 47.78 & 23.33 \\
Qwen2-VL-7B-Instruct & 25.56 & 26.67 \\
GPT-4o & 36.67 & 33.33 \\
GPT-4o mini & 21.11 & 26.67 \\
GPT-4 Turbo & 17.78 & 13.33 \\ \bottomrule[1pt]
\end{tabular}%
}
\end{table}

\section{Prompts for VLMs}
In this section, we provide the designed prompts for the nine BSA tests to receive results from VLMs.

\subsection{MGMP Spatial Visualization Test (SVT)}
\begin{figure}[H]
    \centering
    \includegraphics[width=0.9\linewidth]{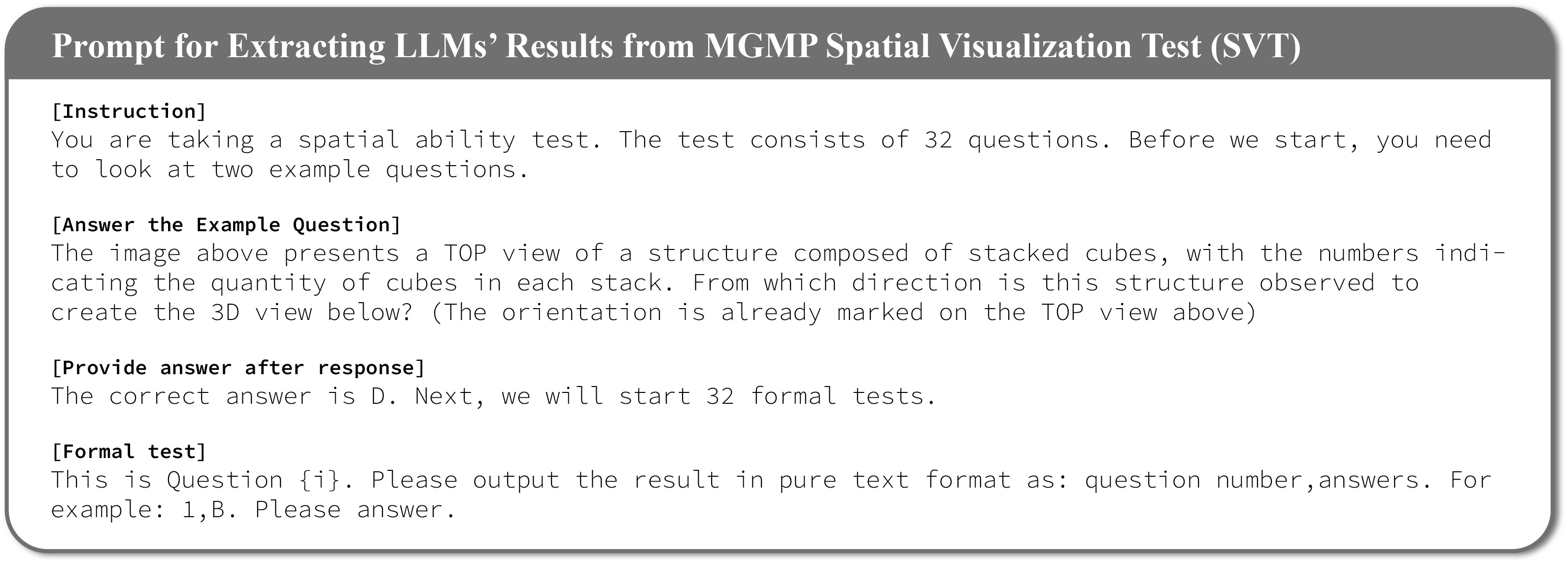}
    \label{fig:prompts_SVT}
\end{figure}

\subsection{Net Cube Imagination Test (NCIT)}
\begin{figure}[H]
    \centering
    \includegraphics[width=0.9\linewidth]{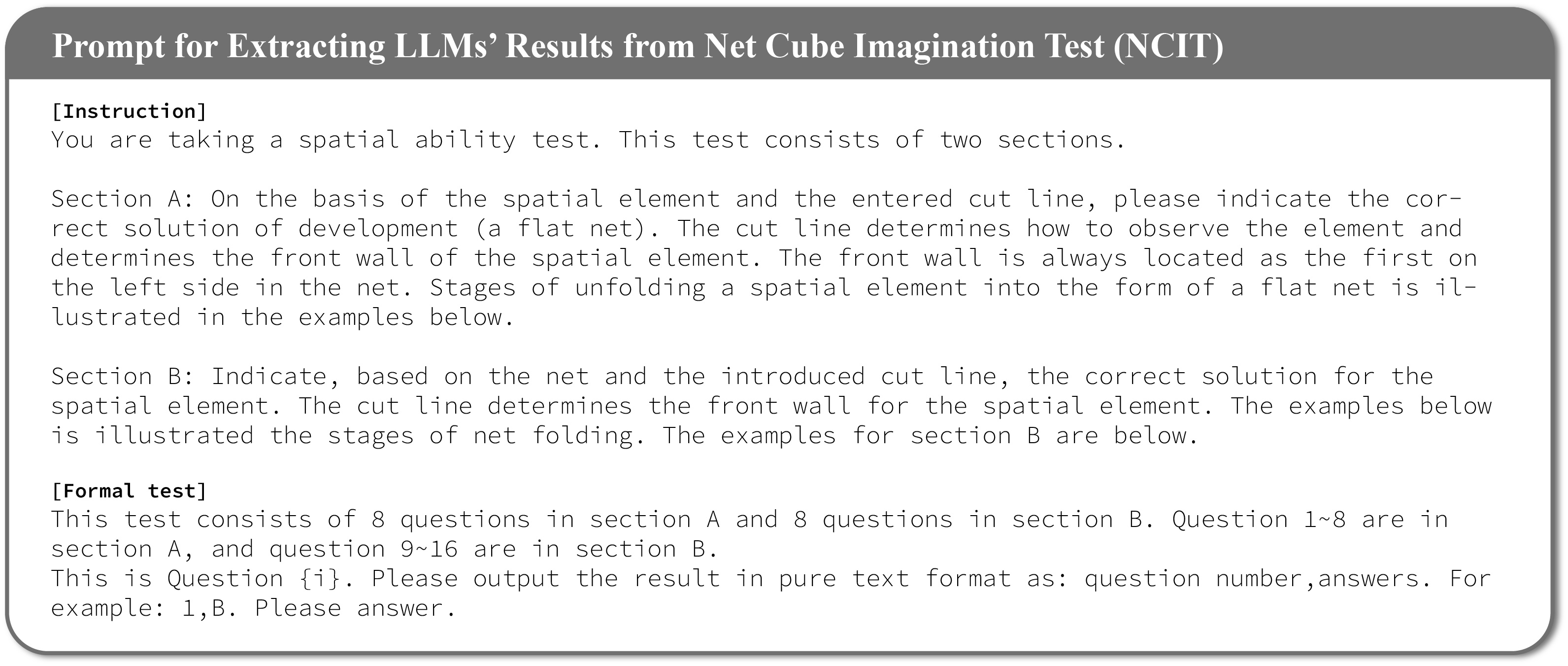}
    \label{fig:prompts_NCIT}
\end{figure}

\subsection{Differential Aptitude Test: Space Relation (DAT:SR)}
\begin{figure}[H]
    \centering
    \includegraphics[width=0.9\linewidth]{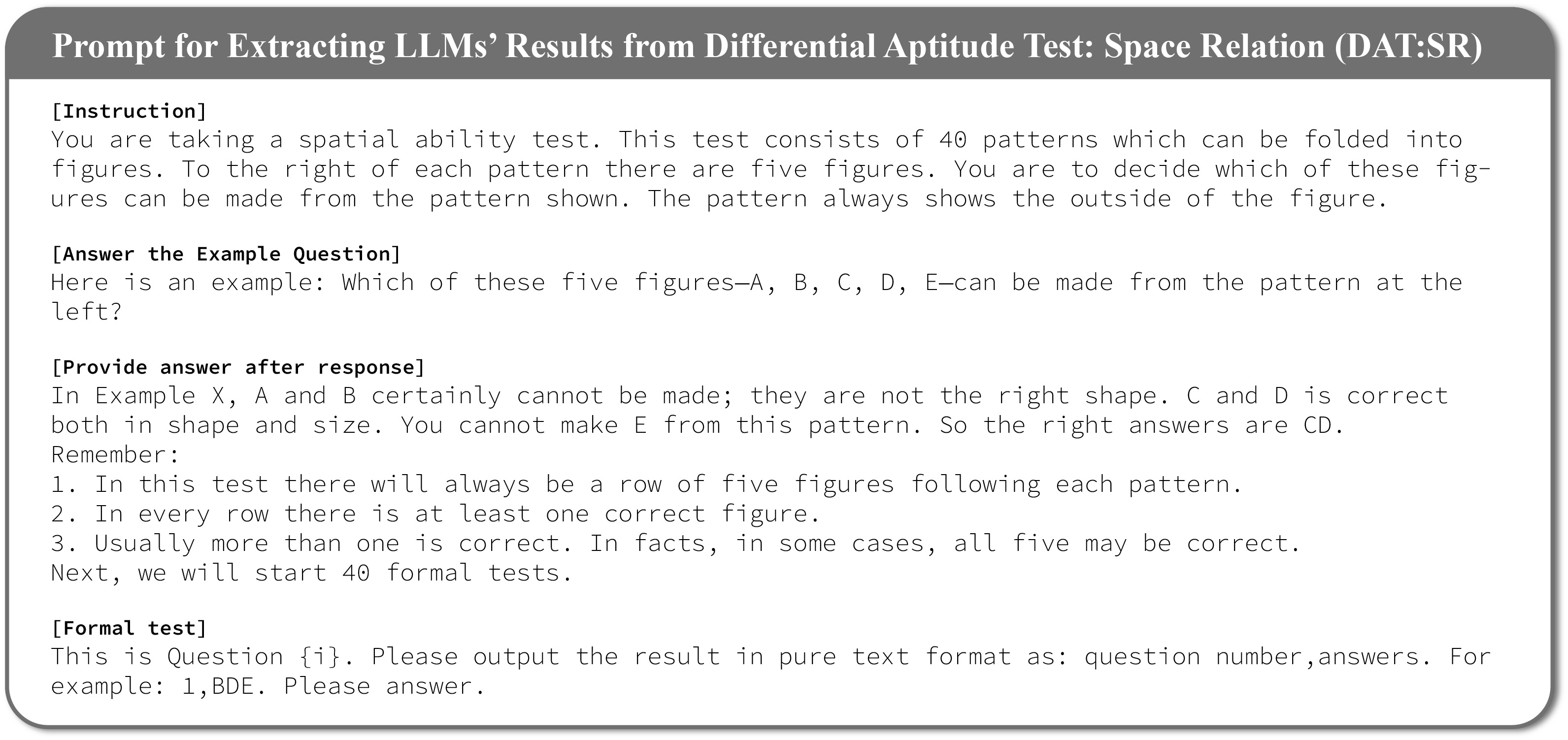}
    \label{fig:prompts_DATSR}
\end{figure}

\subsection{R-Cube-Spatial Relation Test (R-Cube-SR)}
\begin{figure}[H]
    \centering
    \includegraphics[width=0.9\linewidth]{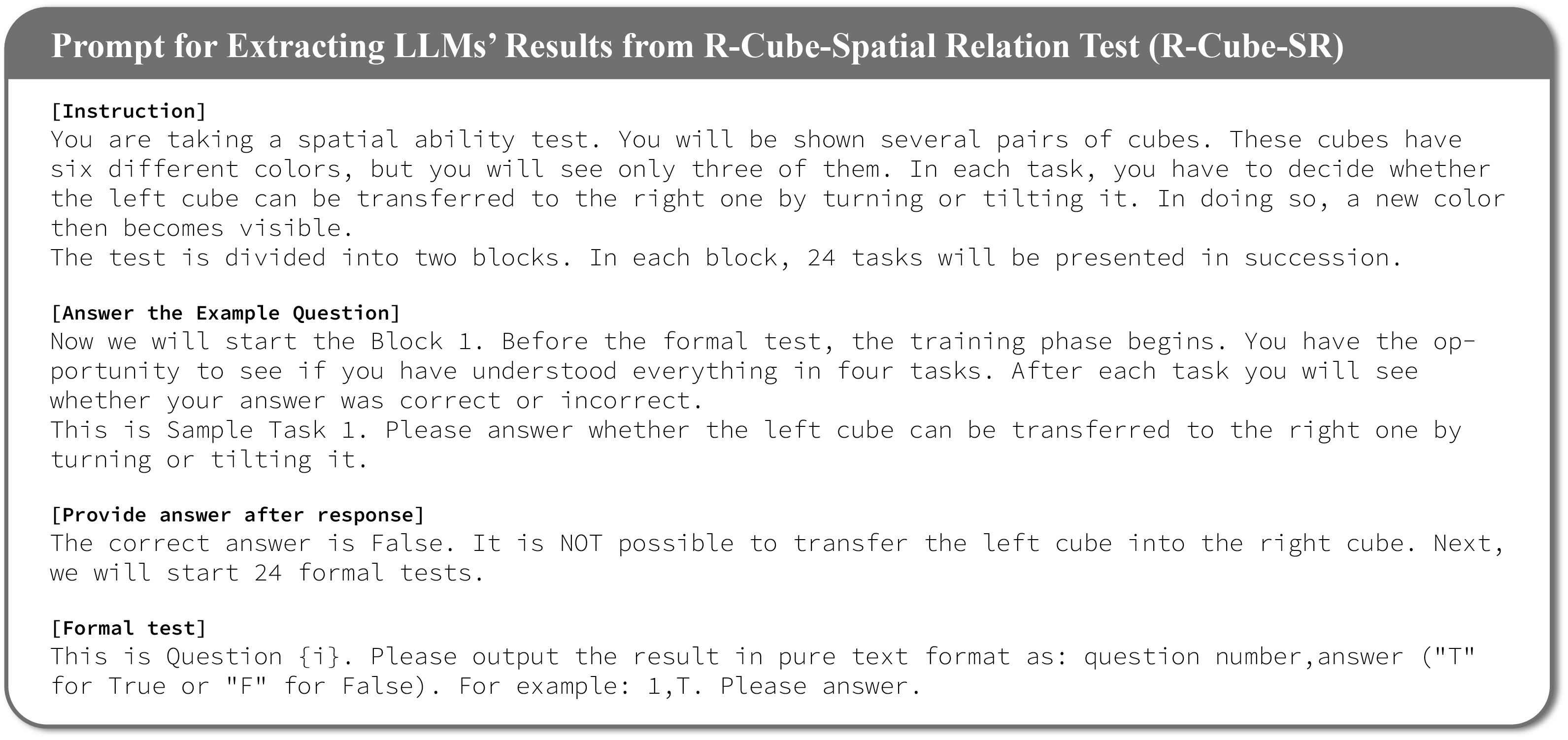}
    \label{fig:prompts_RCUBESR}
\end{figure}

\subsection{Money Road-Map Test (MRMT)}
\begin{figure}[H]
    \centering
    \includegraphics[width=0.9\linewidth]{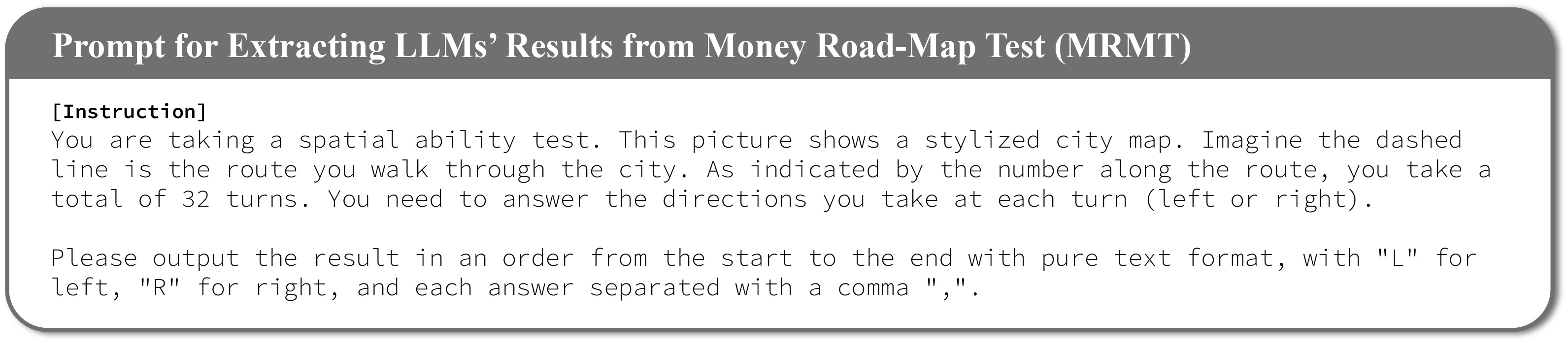}
    \label{fig:prompts_MRMT}
\end{figure}

\subsection{Mental Rotation Test (MRT)}
\begin{figure}[H]
    \centering
    \includegraphics[width=0.9\linewidth]{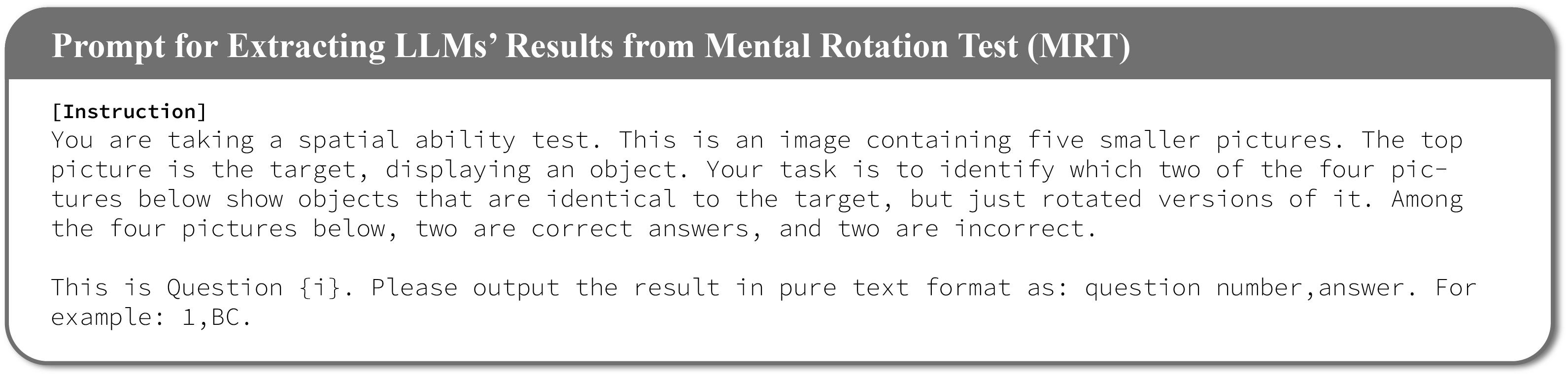}
    \label{fig:prompts_MRT}
\end{figure}

\clearpage

\subsection{Purdue Spatial Visualization Tests: Visualization of Rotations (PSVT:R)}
\begin{figure}[H]
    \centering
    \includegraphics[width=0.9\linewidth]{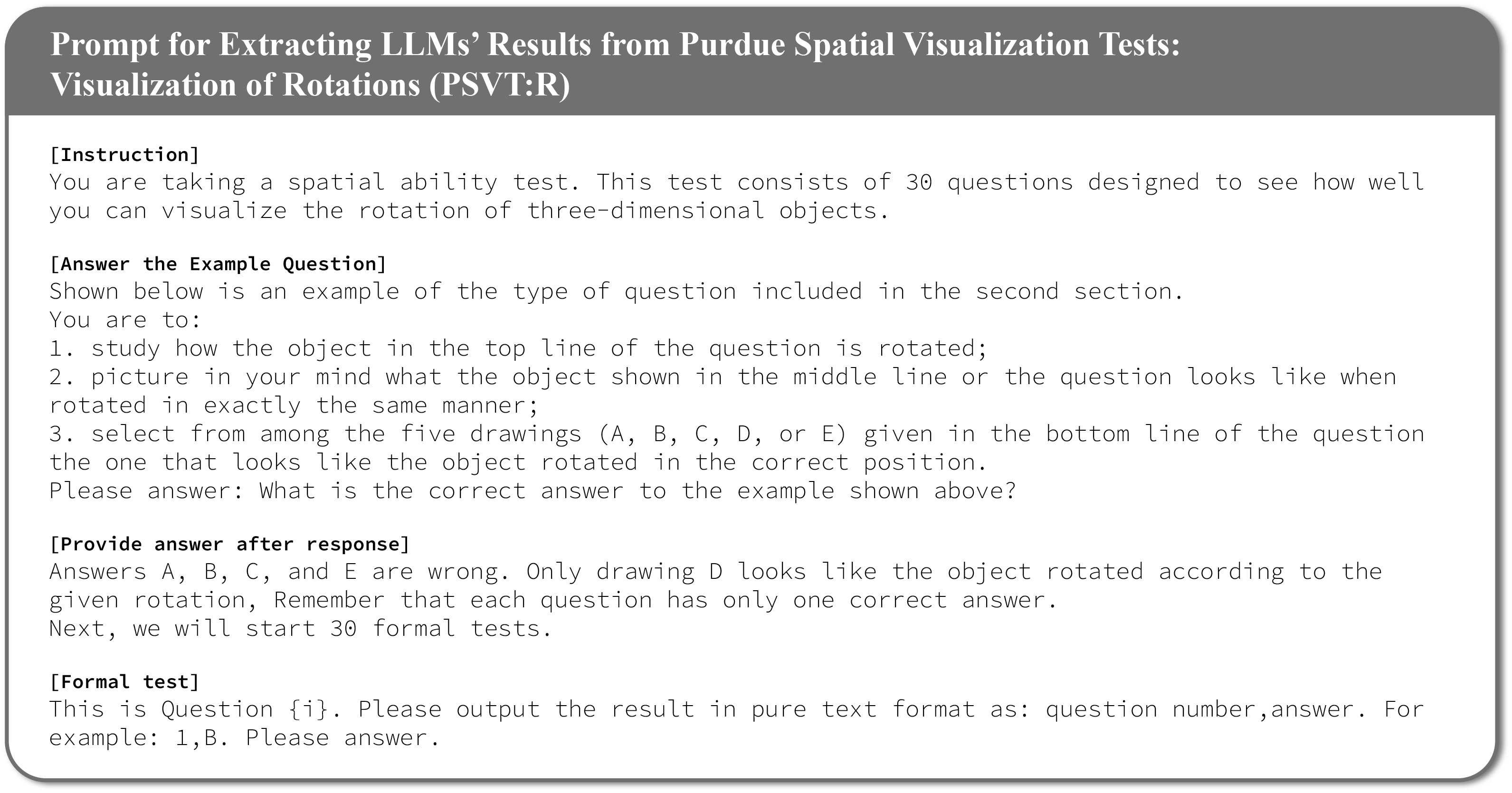}
    \label{fig:prompts_PSVTR}
\end{figure}

\subsection{Santa Barbara Solids Test (SBST)}
\begin{figure}[H]
    \centering
    \includegraphics[width=0.9\linewidth]{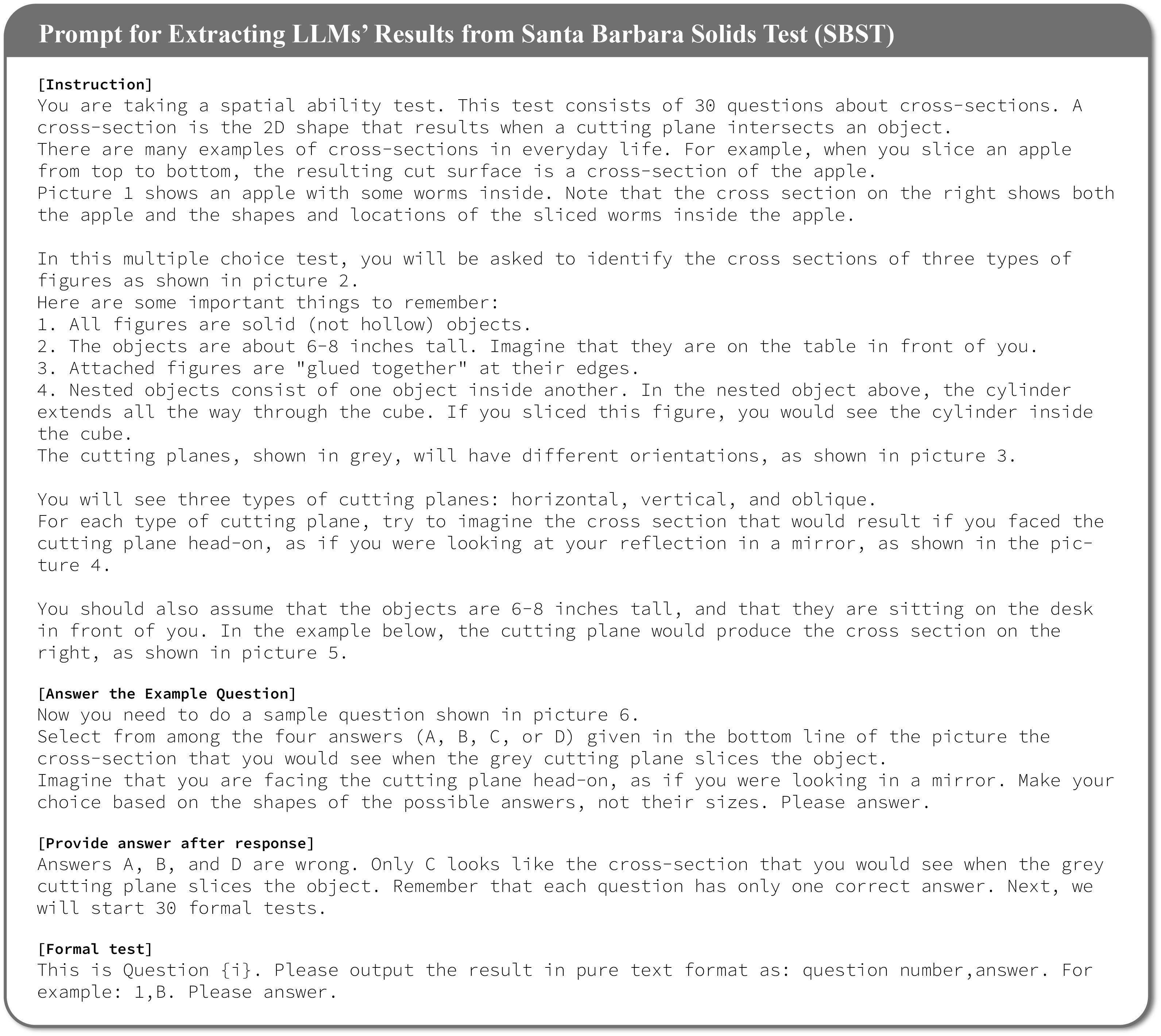}
    \label{fig:prompts_SBST}
\end{figure}

\clearpage

\subsection{R-Cube-Visualization Short Test (R-Cube-Vis)}
\begin{figure}[H]
    \centering
    \includegraphics[width=0.9\linewidth]{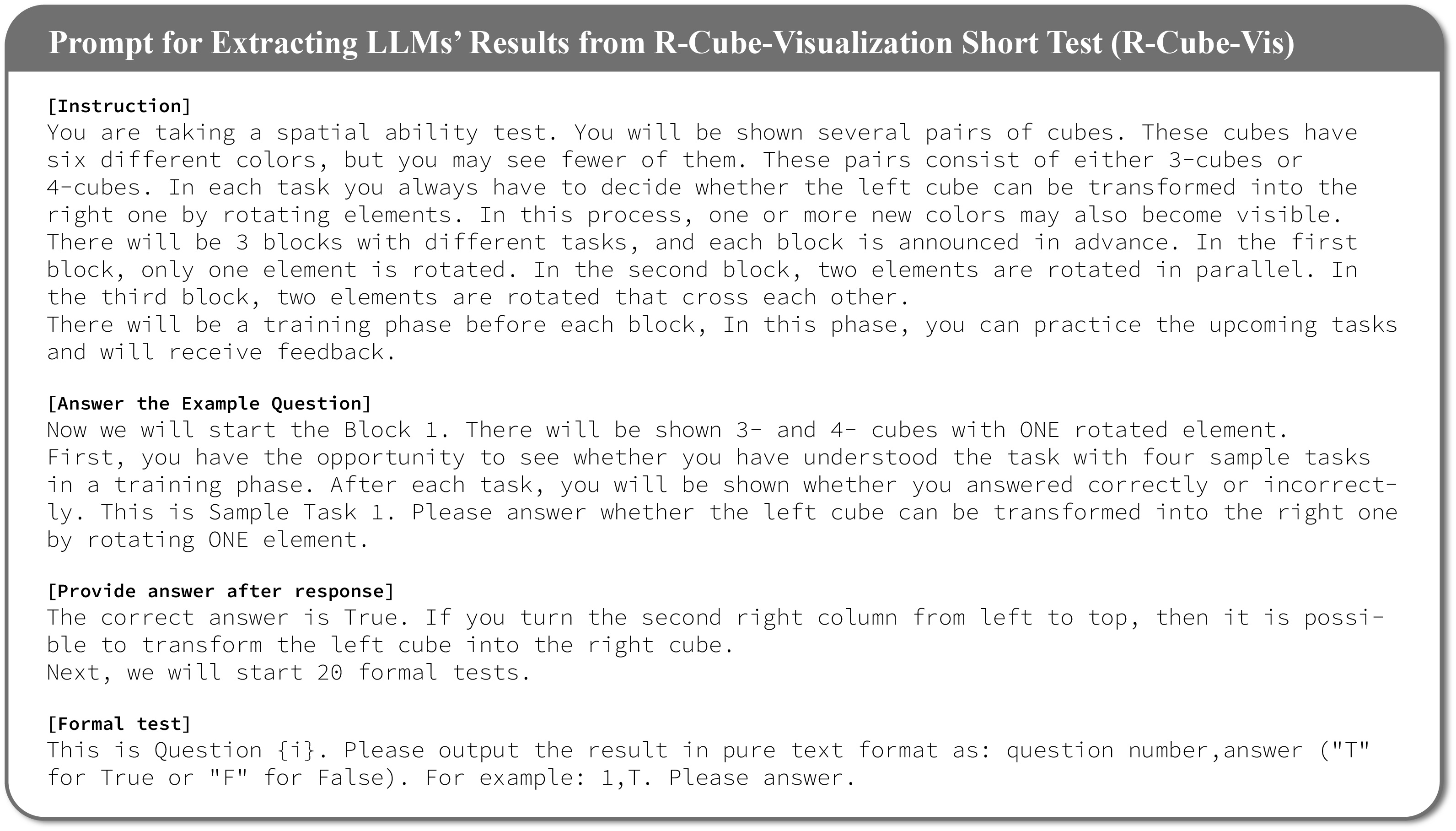}
    \label{fig:RCUBEVIS}
\end{figure}

\section{CoT Method and Example-based Training Demonstration}
In this section, we selected the SBST test, using GPT-4o to evaluate the extent to which the chain-of-thought (CoT) method and example-based training enhance the model's ability to solve spatial problems.

In the first experiment, we designed a structured CoT reasoning process (understanding the 3D shape, analyzing the plane, determining the cross-section, matching cross-section to options, giving the answer) along with a single example question. We then assessed the model's accuracy under four different conditions: (a) directly answering without additional guidance, (b) using only example-based training, (c) employing only the CoT method, and (d) combining both CoT and example-based training, where the example question was analyzed using the CoT approach.

In the second experiment, we examined the impact of varying the number of example questions (1, 3, 5, and 10) on response accuracy, without incorporating the CoT method.

\begin{figure}[H]
    \centering
    \includegraphics[width=0.9\linewidth]{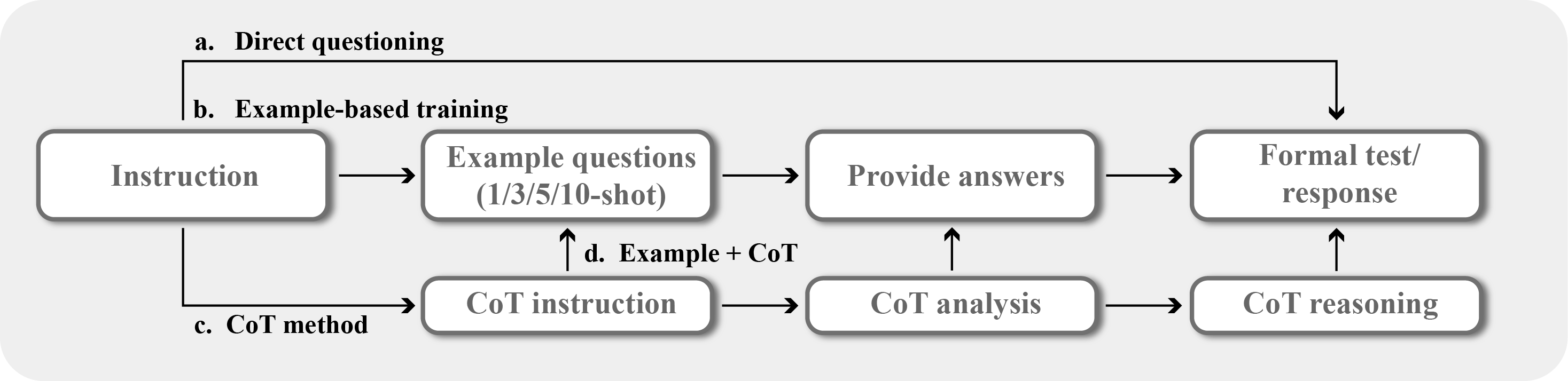}
    \caption{Framework of CoT Method and Example-based Training}
    \label{fig:CoT}
\end{figure}

To enhance the reliability of the findings, both experiments were conducted multiple times to take the average result. The experimental framework is illustrated in Figure \ref{fig:CoT}.

\clearpage

\section{Sample Data Demonstration}
\begin{figure}[H]
    \centering
    \includegraphics[width=0.9\linewidth]{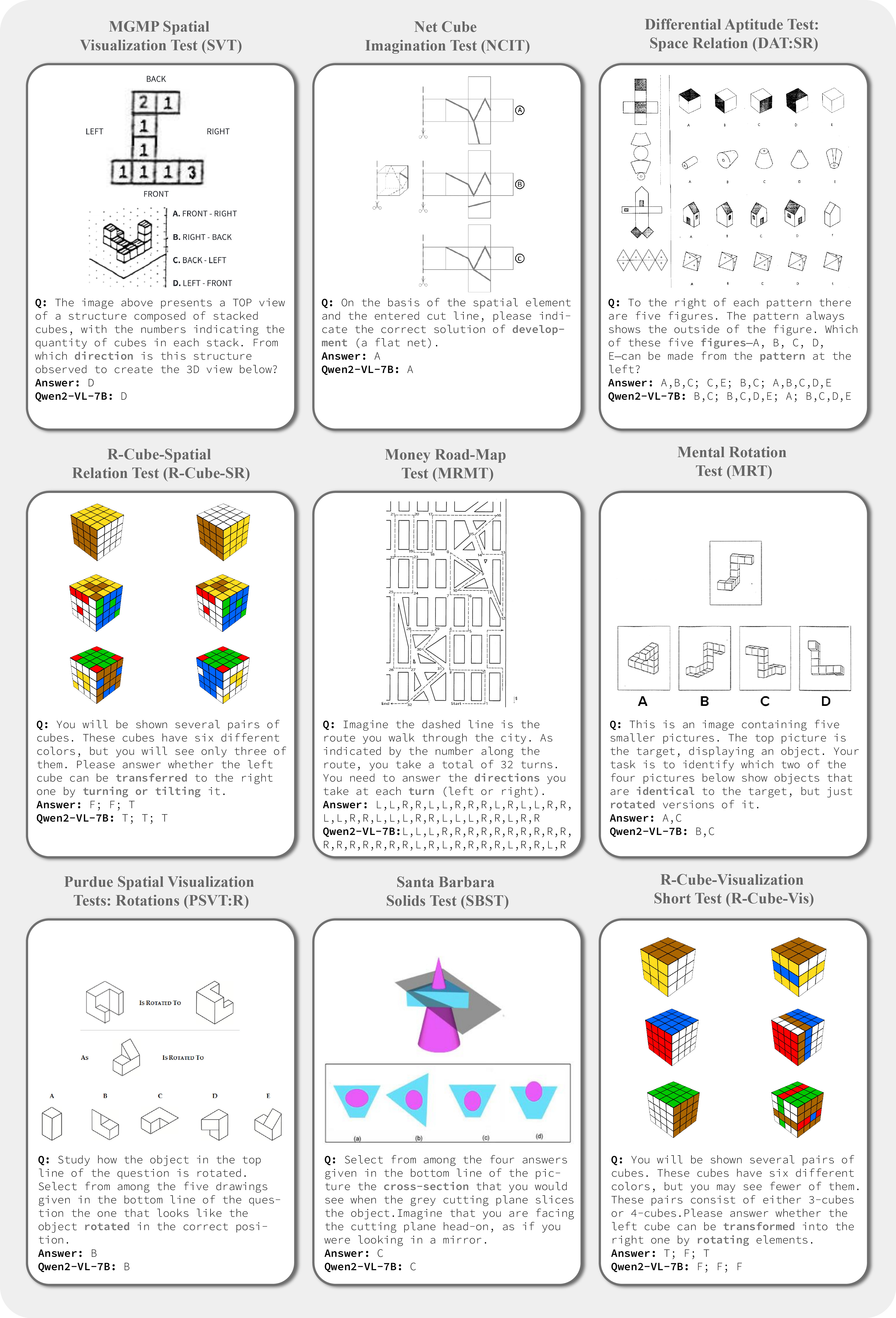}
    \caption{Data samples from tests for BSA\footnotemark[2].}
    \label{fig:Data Samples}
\end{figure}

In this section, we take Qwen2-VL-7B, the model that acquired the highest score as an example, to demonstrate its response to nine sample questions of different tests, as shown in Figure \ref{fig:Data Samples}.

\end{document}